\documentclass[preprint,journal]{vgtc}

    \usepackage{float}

\usepackage[section]{placeins}
    
    \title{\systemname{}: An Interaction Framework for Tuning Text Embeddings Based on Human Feedback}
    
\author{
  \authororcid{Yan Zhu}{0009-0001-3430-5414},
  \authororcid{Y. Chen}{0009-0007-5280-0690},
  and \authororcid{Rebecca Faust}{0000-0002-7640-1287}
}
    
    \authorfooter{
      \item Yan Zhu is with Tulane University.  
            E-mail: yzhu27@tulane.edu.
      \item Y. Chen is with Tulane University.  
            E-mail: ychen88@tulane.edu.
      \item Rebecca Faust\textsuperscript{*} is with Tulane University.  
            E-mail: rfaust1@tulane.edu.
    }

    \abstract{
    In large-scale text analysis tasks, pre-trained language models are often used to embed text corpora for downstream analysis. However, such models may struggle to capture domain-specific semantics and adapting them typically requires large amounts of labeled data and technical expertise to implement training pipelines. Recent approaches have demonstrated how visual interactions in document projections can capture human feedback as training signals for model tuning. However, these methods operate on document-level feedback, which requires users to open and assess individual documents in order to provide effective feedback. In this paper, we propose \systemname{}, an interaction framework that enables feature-level feedback through keyword-based concept specification.  Users specify feedback by organizing extracted keywords into groups representing concepts, which \systemname{} translates into document-level supervision for subsequent tuning. By operating on keywords as the primary interaction medium, \systemname{} reduces the need for manual document inspection and labeling and lowers the barrier to adapting embedding models. We present a prototype implementation that, given a corpus, curates representative keywords, visualizes keywords and document embeddings via dimensionality reduction, allows interactive specification of keyword groups, and supports iterative refinement through system feedback. We evaluate \systemname{} through a user study, usage scenarios, and quantitative experiments demonstrating its effectiveness in capturing user intent and improving embedding alignment.
    }
  
    
    \teaser{
      \centering
      \includegraphics[width=\linewidth, alt={Before and after views from KeySI demonstrating model tuning}]{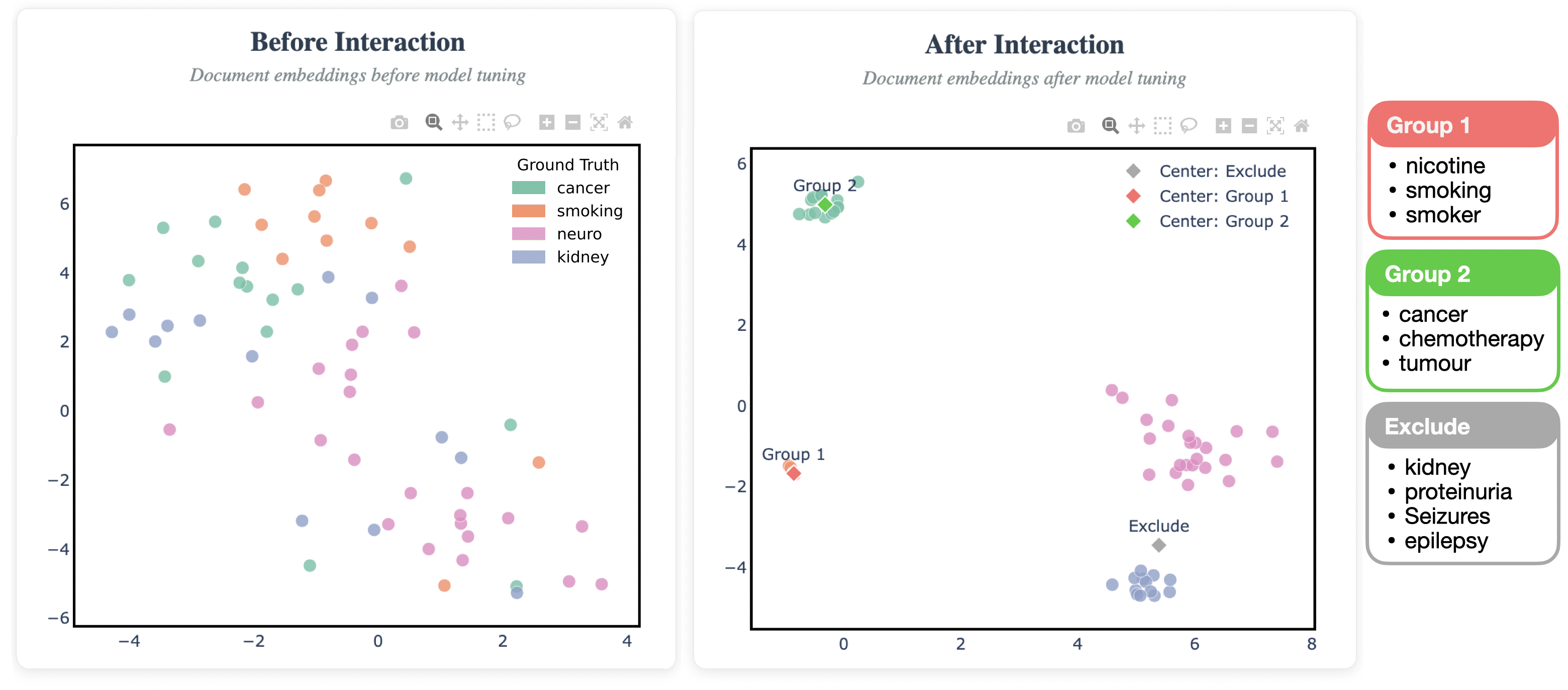}
      \vspace{-2em}
      \caption{ \systemname{} on a dataset of COVID-19 article abstracts containing four risk factors: smoking, cancer, neurological, and kidney. The initial embeddings (left) do not meaningfully capture the risk factors. Using \systemname{}, a user specifies keyword groups to define two of the risk factors (right) and adds out-of-scope terms related to the other two risk factors in the \texttt{Exclude} group. Using this feedback, KeySI tunes the model. The resulting embeddings (center) more clearly capture the two specified risk factors and also reveal better organization of documents related to the remaining factors. Note that the labels and coloring are only used for demonstration.
      }
      \label{fig:teaser}
    }
    \usepackage{xcolor}
    \usepackage{booktabs}
\usepackage{multirow}
\usepackage{adjustbox}

    \graphicspath{{figs/}{figures/}{pictures/}{images/}{./}} 
    \usepackage{amsmath} 
    \usepackage{amssymb}
    \usepackage{tabu}                      
    \usepackage{booktabs}                  
    \usepackage{lipsum}                    
    \usepackage{mwe}                       
    \usepackage{graphicx}
    \usepackage{subcaption}  
    \usepackage{mathptmx}                  
    \usepackage{comment}
    \usepackage{xcolor}
    \definecolor{darkgreen}{rgb}{0,0.5,0}
    
    \newcommand{\systemname}{KeySI}

    \definecolor{steelblue}{HTML}{3C75AF}
    \definecolor{lightorange}{HTML}{E9A888}

\usepackage[dvipsnames]{xcolor}
\usepackage[normalem]{ulem}
\definecolor{mred}{rgb}{.80,.12,.30}
\definecolor{MRED}{rgb}{.80,.12,.30}
\definecolor{grey}{rgb}{0.5,0.5,0.5}
\definecolor{purple}{rgb}{.75,0,.85}
\definecolor{pistachio}{rgb}{0.58, 0.77, 0.45}

\newif\ifnotes

\newcommand{\add}[1]{\ifnotes{\leavevmode\color{mred}{#1}}\else{#1}\fi}

\begin{document}
    
    
\firstsection{Introduction}
    
    \maketitle
    Recent years have seen a dramatic increase in the availability of pre-trained language models~\cite{wang2023pre}, across a variety of domains~\cite{ho2024surveypretrainedlanguagemodels}.
    The emergence of these pre-trained models has resulted in a shift from experts creating small, task-specific language models, to a ``pre-training then fine-tuning'' paradigm~\cite{wang2023pre}. In this paradigm, rather than training from scratch with large amounts of data, users are able to fine-tune pre-trained models with smaller amounts of annotated data. This new paradigm  lowers the barrier to applying language models by reducing both labeled data requirements and the technical effort needed compared with training from scratch. However, despite these advancements, it may still be prohibitively high for non-expert users with little to no technical expertise to implement training pipelines or any users curating large sets of labeled data for tuning~\cite{mishra2021designing}.
More fundamentally, existing supervision mechanisms often operate at the level of labels or individual instances~\cite{bian2021deepsi,mishra2021designing}. This makes it harder to directly express higher-level semantic concepts that may span many documents~\cite{lam2024concept}.
    
     To reduce the reliance on large labeled datasets and technical expertise, recent approaches have begun employing interactive visualization as an interface for pseudo-supervision in DL models~\cite{bian2021deepsi, bian2024neuralsi, lin2024imagesi}. The visualization itself provides a window into the current state of the embedding space, while accompanying interactions allow users to convey domain knowledge through visual operations. These approaches then convert interactions into learnable feedback, thus tuning the model without requiring labeled data or deep technical expertise. For example, Bian and North's DeepSI~\cite{bian2021deepsi} and Lin et al.'s ImageSI~\cite{lin2024imagesi} allow users to interactively tune language and image models (respectively),  relative to the domain information in the data. They rely on  dimension reduction (DR) to visualize the current embedding space and allow direct manipulation of the DR space to convey learnable feedback. 
    
    However, these approaches implicitly assume users know which data items to manipulate to convey feedback. For images, directly plotting the image icons in the DR plot partially overcomes this limitation by cuing users to the contents of the data collection and specific instances to manipulate. For text, however, users must manually inspect individual documents to identify relevant texts for manipulation. This limits the utility of interactions directly in the document space, as manually
    searching through many documents for relevant information requires
    user overhead similar to manually labeling many documents. 
Additionally, document manipulation constrains users to express semantic intent indirectly through representative documents, which may not explicitly capture the desired concept, rather than allowing them to specify concepts directly through explicit signals such as keywords.
    
    

To address these limitations, we propose \systemname{}, a semantic interaction framework for text that shifts the interaction entry point from document-level feedback to feature-level feedback to reduce the manual overheads of document-level interactions and support more explicit concept formation. Specifically, KeySI allows users to specify feedback at the keyword-level by organizing representative keywords into cohesive groups. \systemname{} operationalizes this feedback by translating keyword groups into document-level pseudo-supervision by retrieving relevant documents and organizing them into pseudo-classes, one per user-defined group. These pseudo-classes are then used as labeled data for model tuning. We argue that keyword-level interaction provides a lower-cost yet sufficiently strong supervision signal for steering text embeddings, making it more suitable for early-stage concept specification than document-level interaction. 
    


To evaluate \systemname{}, we perform a three-phase evaluation. First, to demonstrate the use of \systemname{} on real data, we present usage scenarios that illustrate KeySI's ability to integrate user intent into model embeddings. Second, we perform a user study that (1) examines how users employ keyword-level interaction to express semantic intent and adapt text embeddings and (2) compares KeySI's keyword-level interaction to the document-level interaction in DeepSI, demonstrating the reduced manual overhead of keyword-level interactions. Finally, we perform quantitative experiments to evaluate whether keyword-level interaction produces supervision signals strong enough to meaningfully restructure the embedding space.

    
    
    In summary, this paper makes the following contributions:
\begin{enumerate}
    \item \textbf{A keyword-centric semantic interaction framework for text embeddings.}
    We introduce KeySI, a keyword-based semantic interaction approach for model tuning that shifts the interaction entry point from document-level to keyword-level concept specification, reducing the manual overhead.

    \item \textbf{A workflow for translating keyword-level concept feedback into trainable supervision.}
    \add{KeySI integrates and extends existing interaction, retrieval, and embedding-adaptation techniques with a gap-based semantic denoising strategy into a unified workflow for translating keyword-level concept feedback into trainable supervision.}

    \item \textbf{Empirical evidence for the value of keyword-level interaction in text embedding adaptation.}
    Through a user study, usage scenarios, and quantitative analyses, we show that keyword-level interaction reduces the cost of concept specification and improves alignment between embedding structure and user intent.
\end{enumerate}

\section{Related Work}
    
    \subsection{Text Sensemaking}

\add{Text sensemaking aims to help users extract insights and build coherent interpretations from large text collections. Prior visual analytics systems have supported this process at different abstraction. Document- and entity-centered systems, such as Jigsaw and its follow-up studies, help users inspect, relate, and organize individual documents and entities during analysis~\cite{stasko2007jigsaw,kang2012examining,gorg2012combining}.  Other systems support flexible exploration of document collections through human-in-the-loop retrieval and reorganization~\cite{bradel2015big,dowling2019interactive}. These systems demonstrate the value of interactive visual support for
text sensemaking, but largely assume that users express intent
by inspecting, selecting, or manipulating documents.

        \begin{figure}[t]
        \includegraphics[width=\linewidth, alt={Overview of the KeySI workflow. Starting from a text corpus, KeySI extracts representative keywords and document embeddings for visualization. The user organizes keywords into concept groups in the interface. A feedback translator converts the keyword-level interactions into document-level supervision, which is used to fine-tune the embedding model. The updated embeddings are then returned to the visualization for further inspection and iterative refinement.}]{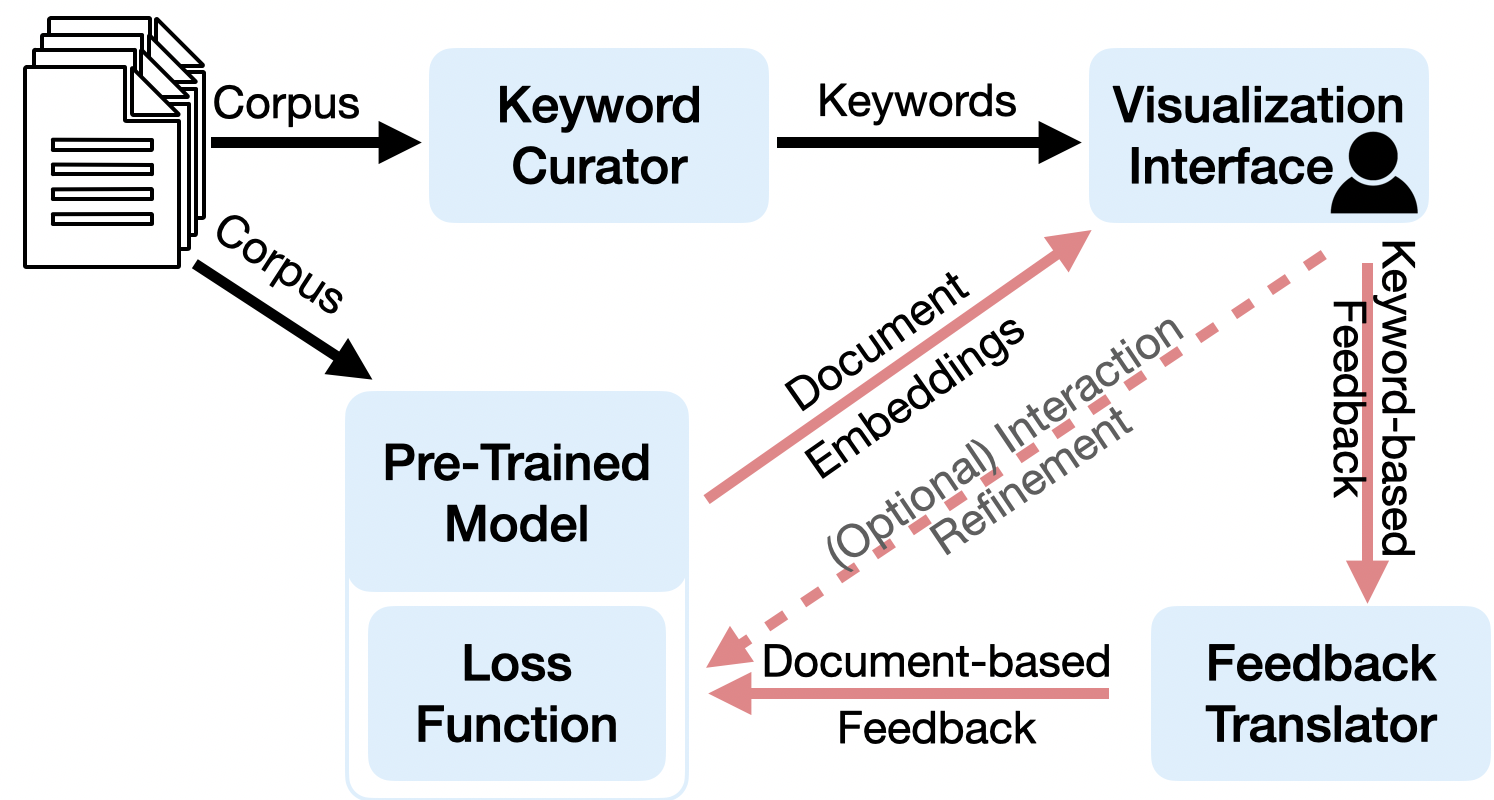}
         \vspace{-1.5em}
        \caption{\textbf{KeySI Workflow.}
        KeySI curates a set of keywords from the corpus to pass to the visualization interface, along with the document embeddings. In the interface, the user constructs keyword-based feedback. This feedback is then passed to the feedback translator which converts it into document-based feedback, making it compatible with the model for tuning. This feedback is integrated into the model and updated embeddings are passed back to the visualization interface for inspection by the user. From here, the user can either refine their initial interaction or provide additional keyword-based feedback. }
        \label{fig:overview}
         \vspace{-1.5em}
    \end{figure}


A related line of work supports sensemaking through topic-, lexicon-, or concept-level interaction. Tiara leverages topic modeling to enable interactive visual summarization~\cite{liu2009interactive}. Utopian enables users to steer topic models through feedback on topics and terms~\cite{choo2013utopian}; ConceptVector supports interactive lexicon construction using word embeddings~\cite{park2017conceptvector}; and Semantic Concept Spaces uses word-embedding projections to guide topic-model refinement~\cite{elassady2019semantic}. These systems reduce the need to operate only on individual documents by allowing users to work with higher-level semantic constructs. However, their primary goal is still to support exploration, topic refinement, or lexicon construction, rather than translating user-defined concepts into supervision for adapting a document embedding model. 

In contrast, KeySI shifts the interaction from manipulation of individual documents to keyword-based concept specification, only requiring document interaction for verification and refinement. Additionally, rather than focusing on topic modeling, KeySI translates user-defined concepts into pseudo-supervision to adapt the underlying document embedding model.}



    \add{\subsection{Human Feedback and Representation Adaptation}}
    
    \add{Prior work has explored the role of embeddings in visual analytics and human-centered analysis workflows~\cite{huang2023va}. It has examined how human feedback and semantic knowledge can improve text analysis models and adapt representation spaces. One line of research focuses on task-specific adaptation. For example, Sakai and Masuyama showed that keyword-level feedback can recalibrate multi-document summarization~\cite{sakai2004multiple}. SemanticPush translates user intent into synthetic training examples to steer text classifiers~\cite{kiefer2022semantic}, while other methods support more direct forms of user control over learned representations through manipulating learned predicates, iteratively refining semantic rules, or interactive transfer learning ~\cite{sen2019heidl,yang2019study,mishra2021designing}. These approaches demonstrate the value of human feedback, but are typically tied to specific tasks or explicit rule/label structures.

    \begin{figure*}[t]
        \centering
        \includegraphics[width=.95\textwidth, alt={Screenshot of the KeySI user interface. The interface consists of six labeled panels. Panel A provides controls for creating keyword groups and starting model tuning. Panel B displays extracted keywords for browsing and selection. Panel C shows a two-dimensional document embedding projection where documents related to the selected keyword or concept group are highlighted. Panel D lists user-created keyword groups together with the Exclude group. Panel E displays documents associated with the selected keyword or group and their top five keywords. Panel F shows the full text of the selected document. Colors distinguish keyword groups, purple stars indicate documents associated with the active keyword or group, and a red star marks the currently selected document.}]{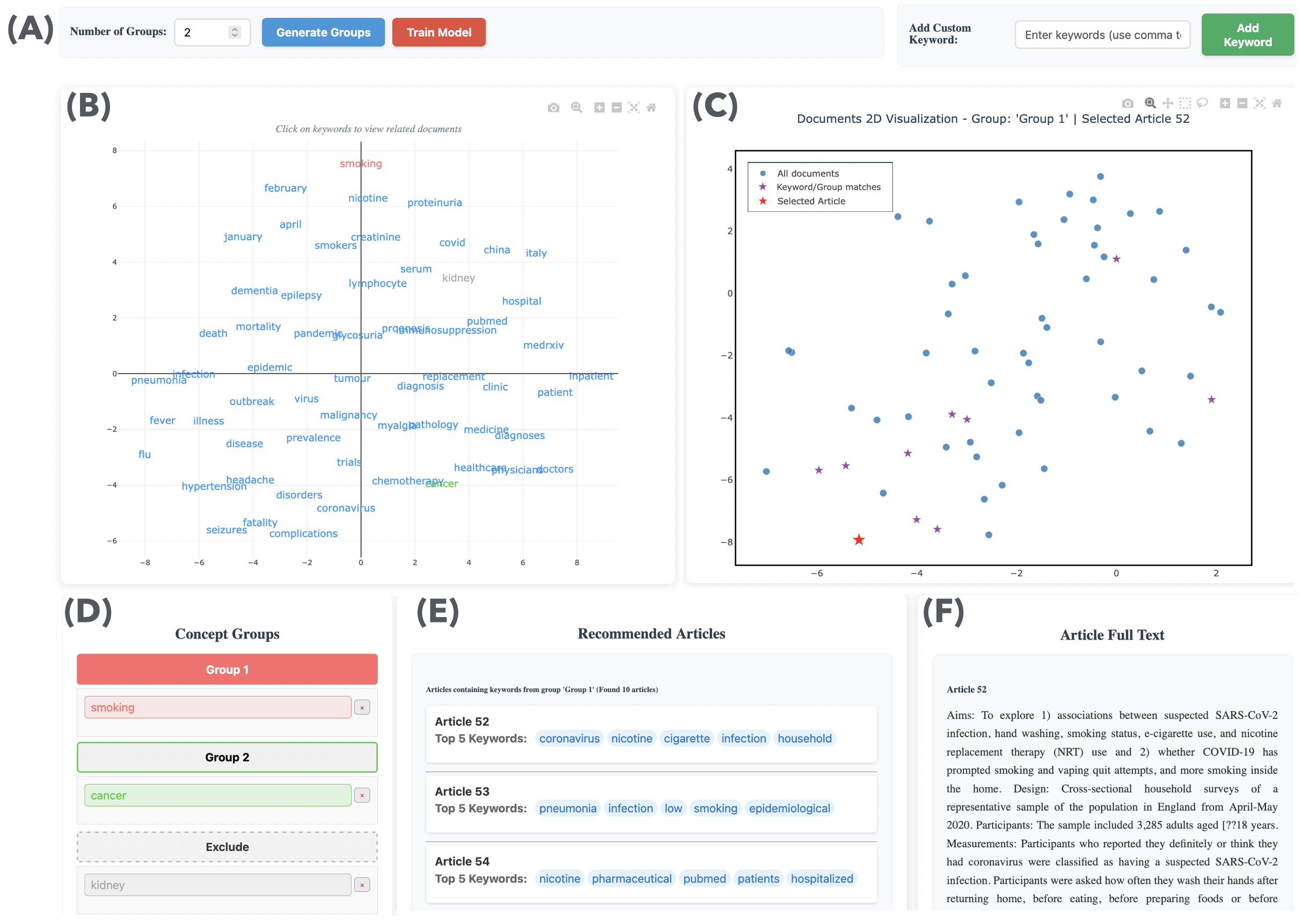}  
         \vspace{-1em}
\caption{\textbf{\systemname{} UI.}
\add{(A) Controls for setting the number of keyword groups and starting model tuning.(B) A keyword view for browsing and selecting extracted keywords.(C) A 2D document projection in which documents associated with the selected keyword or concept group are highlighted.(D) A group panel showing user-created keyword groups and the \texttt{Exclude} group.(E) A document list showing documents associated with the selected keyword or group together with their top-5 keywords.(F) A full-text view of the selected document.
Colors indicate user-defined keyword groups; purple stars mark documents associated with the active keyword or group, and the red star marks the currently selected document.}}
        \label{fig:UI}
        \vspace{-2em}
    \end{figure*}

A second line investigates reshaping and inspecting representation spaces. Retrofitting adjusts pretrained word embeddings using structured semantic lexicons~\cite{faruqui2015retrofitting}, while visualization has been used to explain embedding alignment processes such as semi-supervised text alignment~\cite{meinecke2021explaining}.  These works highlight the potential of embedding adaptation and inspection, yet they typically rely on predefined lexical resources, offline constraints, or analysis tools rather than lightweight, interactive concept-level feedback.

A third line employs interactive clustering, where users guide document organization through feature-level feedback or visual refinement. Some approaches allow users to label, select, or reweight informative terms to steer clustering~\cite{bekkerman2007interactive,nourashrafeddin2013interactive,hu2011interactive,hu2014interactive}, while others provide visual analytics interfaces for inspecting and refining clusters or clustering workflows~\cite{lee2012ivisclustering,sherkat2018interactive,sherkat2019visual,rezaeipourfarsangi2022interactive,ruppert2017visual}. 
Although these systems support document grouping and cluster refinement, their primary objective is typically to improve the organization of a corpus rather than to adapt an underlying document embedding model.

KeySI builds on these directions but differs in both interaction form and adaptation target. Rather than depending on labeled examples, explicit rules, predefined lexicons, or direct clustering refinement, KeySI derives supervision from user-defined keyword-based concepts and translates them into document-level pseudo-supervision to steer the embedding space toward user-specified semantic distinctions.}
 
\subsection{Semantic Interaction}
\add{Semantic interaction (SI) describes a paradigm in which users communicate analytical intent through visual interactions that dynamically influence the underlying computational model~\cite{endert2012semantic}. Recent work has extended SI to deep representations and embedding-based visual analytics. DeepVA~\cite{bian2019deepva} integrates visual interactions by re-weighting deep features, while DeepSI~\cite{bian2021deepsi} and ImageSI~\cite{lin2024imagesi} allow users to steer text and image embedding models by directly manipulating projected instances in the visualization. These approaches show how visual interaction can align learned representations with user intent, but they primarily rely on instance-level manipulation. For text collections, this often requires users to locate, inspect, and spatially manipulate individual documents before providing effective feedback.

KeySI adopts the semantic-interaction paradigm but shifts the primary interaction object from projected documents to keyword-defined concepts. Document views are retained for verification and refinement, allowing users to inspect concept assignments without making document manipulation the main feedback mechanism.}

    \section{\systemname{}}
    A core contribution of this work is a concept-level interaction mechanism that enables users to express semantic intent and translate it into learnable supervision for model steering. \add{Figure~\ref{fig:overview} presents the workflow of this framework.} \add{The key idea is that users provide feedback through keyword-based concept specification, formed through interactions with keyword visualizations.} From this feedback, the system infers user intent and translates it into document-level supervision for model tuning. \add{Users then progressively validate and revise them through linked document inspection, model updates, and optional refinement within a human-in-the-loop workflow.} This framework allows users to teach the model broader semantic concepts through feature-level interactions without requiring labeled data or exhaustive manual inspection of documents.

    We present \systemname{}, our implementation of this framework. Section~\ref{sec:Interaction workflow} introduces the \systemname{} interaction workflow, Section~\ref{sec:system} presents the system pipeline, Section~\ref{sec:model-tuning} presents the tuning methodology, and Section~\ref{sec:refinement} describes an additional refinement interaction. 
    
\subsection{Interaction Workflow}
    \label{sec:Interaction workflow}


\add{KeySI elicits user feedback through the interface shown in Fig.~\ref{fig:UI}. After loading a corpus, KeySI performs \textbf{keyword extraction and curation} and organizes representative keywords into the keyword view for selection (Fig.~\ref{fig:UI} (B)), providing an initial overview of the corpus. Users may observe clusters of related keywords, infer preliminary concepts represented in the document collection, and select keywords that match those emerging concepts. To specify a concept, users first select an active concept group in the group panel (Fig.~\ref{fig:UI} (D)) to add words to and then click keywords in the keyword view (Fig.~\ref{fig:UI} (B)) to add them to that group. Keyword colors in (B) indicate their assigned user-defined groups in (D).

The document projection (Fig.~\ref{fig:UI} (C)) provides structural context for how the current model represents the corpus. When users select a keyword in (B) or concept group in (D), linked interactions highlight the associated documents in the projection and update the candidate list (Fig.~\ref{fig:UI} (E)) with retrieved documents and keyword previews. This allows users to quickly assess whether a selected keyword or concept group retrieves documents consistent with their intended concept. Users can select documents either from the projection or from the candidate list to inspect their full text (Fig.~\ref{fig:UI} (F)) for ambiguous cases. Users may also define a group of words that are irrelevant to their desired concepts, called the \texttt{Exclude} group (Fig.~\ref{fig:UI} (D), bottom). Once satisfied with their concept groups, users click \textit{Train Model} to pass the keyword-level feedback to the backend for model tuning.}

 In the back end, KeySI \textbf{translates these keyword groups into document-level pseudo-supervision} by \textbf{retrieving related documents} and applying \textbf{semantic denoising}. The resulting document sets are then used to tune the embedding model. After tuning, KeySI presents updated document projections alongside the original view so that users can compare the result, revise keyword groups, or enter a refinement stage to further verify and correct the model update.

    \subsection{KeySI System Pipeline}
    \label{sec:system}
\begin{figure}[t]
    \includegraphics[width=\linewidth, alt={Overview of the KeySI system pipeline. The pipeline consists of four stages connected from left to right. KeySI first extracts representative keywords from the text corpus. Users then interact with visualizations of keywords and document embeddings to define concept groups. A feedback translation module converts keyword-level interactions into document-level supervision, which is used to fine-tune the embedding model. The translation and tuning process is performed twice: first using a strict translation for initial model adjustment, and then using a relaxed translation based on the updated model.}]{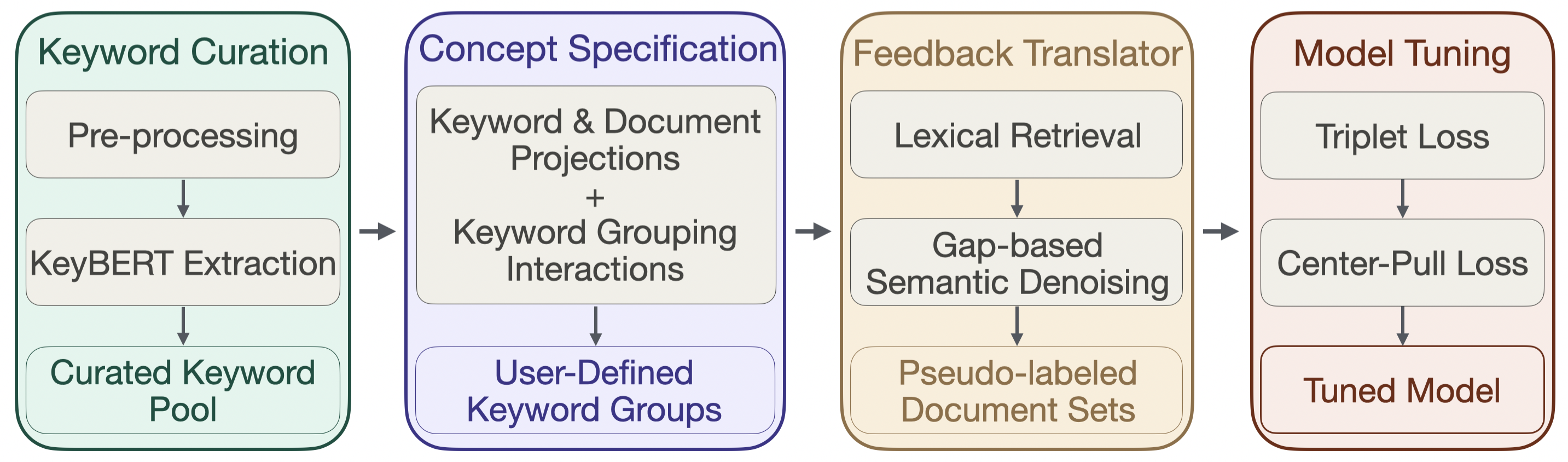}
         \vspace{-1.5em}
        \caption{\textbf{KeySI System Pipeline.} KeySI first curates a set of keywords, then presents visualization of keywords and documents for concept specification. Next, it translates keyword-level feedback into document-level feedback and uses this feedback to tune the model. Note that it performs these last two steps twice - once with a strict translation for an initial model adjustment, and again with a more relaxed translation (using the adjusted model). }
        \label{fig:system}
         \vspace{-1.5em}
    \end{figure}

    This section presents the KeySI system pipeline for the interaction workflow described above, shown in Fig.~\ref{fig:system}.  
    \subsubsection{Keyword Extraction and Curation} KeySI uses keywords as the primary interface for concept specification. It extracts candidate keywords from the corpus and presents them to users for grouping or exclusion, providing an entry point for expressing semantic intent.

    \paragraph{Pre-processing and Keyword Extraction}
    KeySI applies lightweight text pre-processing to support keyword extraction and later retrieval.
    Documents are first cleaned by removing unwanted characters and digits. We then tokenize the text and
    apply part-of-speech tagging using NLTK~\cite{bird2006nltk}, retaining only nouns, as they typically
    represent semantically informative concepts such as entities, topics, and objects. To reduce lexical
    variation and consolidate related terms, we apply Snowball stemming~\cite{bird2006nltk},
    mapping words such as \emph{smoke}, \emph{smoking}, and \emph{smoker} to a shared stem.
    
    Based on the pre-processed text, KeySI extracts representative keywords from each document using
    KeyBERT~\cite{grootendorst2020keybert}. KeyBERT ranks candidate terms by semantic similarity to the
    document embedding, allowing keywords to be selected based on meaning rather than surface frequency.
    We aggregate extracted keywords across the corpus to form the curated keyword pool shown in the interface.
    
    \subsubsection{Concept Specification}
    The curated keywords are projected onto a two-dimensional layout using t-SNE~\cite{maaten2008visualizing} to support early-stage sensemaking. By inspecting them, users can identify terms that reflect their domain concepts and form keyword groups without reading individual documents. 
    In parallel, \systemname{} provides a 2D projection of documents, computed via t-SNE on the pretrained embeddings, as a structural reference for the current model space. Linked interactions connect the keyword and document views, allowing users to assess a concept's scope and coverage before passing the keyword groups to the back end for tuning.

    \subsubsection{Translating Keyword Groups into Learnable Feedback}
    \label{translator}
    To bridge the gap between keyword groups and document-level feedback,
    KeySI translates user-defined keyword groups into pseudo-labeled document sets
    for tuning. A naive approach would create document sets from direct lexical matches to keywords; 
    however, term-based retrieval is only a coarse proxy for semantic
    intent and is prone to semantic ambiguity. Even when a keyword group is semantically coherent, such as {hockey, playoffs, baseball} for a sports concept, one of its keywords may still appear in documents for reasons unrelated to that concept. For instance, \textit{baseball} may retrieve crime news describing ``vandalism with a baseball bat''.
 Such documents satisfy the lexical constraint
    but diverge semantically from the user’s intent. If used as pseudo-supervision, these off-topic matches inject noise, blur concept boundaries, and destabilize fine-tuning.
    
    Thus, KeySI employs a multi-step pipeline that retrieves a candidate pool via lexical matching and then refines this pool using embedding-based semantic denoising. The goal is not to improve retrieval per se, but to reduce noise and produce pseudo-supervision that is more reliable for model updating yet broad enough to reflect the user's concept.
    

    \paragraph{Candidate Retrieval via Stemmed Lexical Matching}
    After users define keyword groups, KeySI retrieves an initial candidate set for each group through stemmed lexical matching. Retrieval is performed on the stemmed text produced in the pre-processing step, so documents containing morphological variants of a keyword are matched consistently. This stage favors recall and produces a broad pool of potentially relevant documents, which is then refined through semantic denoising.

    \paragraph{Gap-based Semantic Denoising}
    
    KeySI treats the pretrained embedding space as a weak semantic prior: although it may not perfectly align with human judgments, it often preserves coarse topical structure such that concept-consistent documents tend to lie closer to their intended concept than off-topic matches. We exploit this property to reduce semantic ambiguity introduced by lexical retrieval and to produce pseudo-supervision that is stable enough for tuning.

    KeySI filters candidates using a gap confidence score computed in the current embedding space by comparing each document against concept group ``prototypes'' to ensure that it is most aligned with the concept group that retrieved it. Prototypes serve as representative embeddings for the entire concept, computed by averaging the centroids of the document sets retrieved for each keyword. This gives each keyword equal influence when forming the group prototype, rather than allowing a single high-frequency keyword to dominate the group representation. The gap score is then computed based on the document's affinity to concept-group prototypes in the current embedding space. 


Formally, the gap score, $G_i$, for each candidate document $d_i$, is computed as follows. Let $\mathbf z_i$ be the L2-normalized embedding of document $d_i$ and $C_k$ be a concept group $k$ containing keywords $w$, $D_k$ be the set of candidate documents retrieved for $C_k$, and $D_{k,w}$ be the candidate documents retrieved for only keyword $w$.

For each keyword $w \in C_k$, we first compute a centroid over its retrieved documents:

\begin{equation}
\boldsymbol{\mu}_{k,w}=
\frac{1}{|D_{k,w}|}\sum_{d_i \in D_{k,w}} \mathbf z_i
\end{equation}

We then define the prototype for concept group $C_k$ as the normalized average of these keyword-specific centroids:

\begin{equation}
\label{eq:proto}
\bar{\boldsymbol{\mu}}_k=
\frac{1}{|C_k|}\sum_{w \in C_k} \boldsymbol{\mu}_{k,w},
\qquad
\mathbf p_k=\bar{\boldsymbol{\mu}}_k/\|\bar{\boldsymbol{\mu}}_k\|_2.
\end{equation}

Next, we define the affinity of each document to a concept group as the cosine similarity between the document embedding $\mathbf z_i$ and the group prototype $\mathbf p_k$:
\begin{equation}
S_{i,k}=\mathbf z_i^\top \mathbf p_k.
\end{equation}

For each candidate document $d_i$, let $C_l$ be the concept group with the highest affinity and let $C_m$ be the concept group with the second-highest affinity. We define the gap score $G_i$ as:
\begin{equation}
G_i = S_{i,l} - S_{i,m}.
\end{equation}

A large $G_i$ indicates confident membership to $C_l$, whereas a small $G_i$ suggests ambiguous membership near a boundary between groups.

Using this score, we filter the candidate documents for each group to obtain cohesive document sets for tuning. For each group $C_k$, we retain only candidates that both (i) belong to the candidate set of $C_k$, (ii) are most strongly associated with $C_k$ in the embedding space, and (iii) exceed a group-specific gap threshold $\tau_k$:
\begin{equation}
\{ d_i \mid d_i \in D_k,\ C_l = C_k,\ G_i \ge \tau_k \}
\end{equation}
where $\tau_k$ is estimated adaptively from the gap-score distribution of candidates in $D_k$ that currently prefer $C_k$.

The \texttt{Exclude} group does not provide positive supervision and is therefore not treated as a learnable concept during tuning. Instead, it serves as an out-of-scope reference in gap-based filtering. We compute an \texttt{Exclude} prototype $\mathbf{p}_{\mathrm{ex}}$ using the same procedure as Eq.~\ref{eq:proto}, and include it in the top-1/top-2 affinity ranking only when its matched documents form a sufficiently coherent direction. This coherence is measured by the concentration of the \texttt{Exclude} centroid, $c_{\mathrm{ex}}=\|\boldsymbol{\mu}_{\mathrm{ex}}\|_2$, with participation in the ranking only when $c_{\mathrm{ex}} \ge \tau_{\mathrm{ex}}$. We further adopt a staged pseudo-supervision strategy that begins with a stricter denoising to create a high-confidence document set, performs initial tuning, and then applies a relaxed denoising to recover additional concept-consistent documents, before a final tuning. Additional details of the staged denoising procedure are provided in the supplemental material.

\subsection{Model Tuning for Visual Semantic Alignment}
\label{sec:model-tuning}
      
    Using the pseudo-labeled document sets produced during feedback translation, KeySI fine-tunes the text encoder to better align the high-dimensional embedding space with the user's intended concepts. Our goal is not to train a general-purpose classifier, but to incorporate user-provided semantic feedback into the learned representation so that documents associated with the same concept become more cohesive, competing concepts become better separated, and ambiguous boundary cases become easier to distinguish.

These updates are learned in the embedding space and can later be inspected through before--after 2D projection views. To support semantic alignment in the embedding space, KeySI uses a staged optimization strategy that jointly improves between-group separation and within-group cohesion. Concretely, KeySI first optimizes a semi-hard triplet loss to enlarge inter-group margins by repelling confusing negatives, and then optimizes a prototype center-pull loss to compact documents within each concept group around a stable group prototype.

    \subsubsection{Semi-hard Triplet Loss}
    \label{sec:triplet}
    For the first stage, we employ triplet loss to separate user-defined concepts in the embedding space. 
    Let $z_i$ denote the L2-normalized embedding of document $i$
    ($\|z\|_2=1$). We construct triplets $(a,p,n)$ where the anchor $a$ and positive
    $p$ come from the same keyword group, and the negative $n$ comes from a
    different group (including \texttt{Exclude} as a negative pool). Let
    $d_{ap}=\|z_a-z_p\|_2$ and $d_{an}=\|z_a-z_n\|_2$. The triplet margin loss is
    \begin{equation}
    \mathcal{L}_{\text{triplet}}=\max\bigl(0,\ d_{ap}-d_{an}+m\bigr),
    \end{equation}
    where $m$ is the margin.
    We prefer \emph{semi-hard} negatives satisfying $d_{ap}<d_{an}<d_{ap}+m$ to focus
    updates on ambiguous boundary cases; otherwise we fall back to the closest
    available negative.
    
   \subsubsection{Prototype Center-Pull Loss}
\label{sec:center-pull}
In the second stage, we perform a center-pull loss to improve within-group cohesion in the embedding space. To do so, we maintain a unit-norm prototype
$\mu_k$ for each concept group $C_k$ in the interaction $I$ (not including \texttt{Exclude}).
Given a training batch, let $B_k$ be the set of documents in the batch assigned to concept group $C_k$. We minimize
the L2 distance to the group prototype:
\begin{equation}
\mathcal{L}_{\text{center}}=
\sum_{C_k\in I}\frac{1}{|B_k|}\sum_{i\in B_k}\|z_i-\mu_k\|_2.
\end{equation}

\subsection{Interaction Refinement}
\label{sec:refinement}
After an initial tuning round, users may still observe mixed regions or ambiguous boundary cases in the updated embedding space. Such discrepancies can arise when retrieved documents only partially match the intended concepts or when the current embedding space doesn't yet fully reflect them. To address this, KeySI provides an interaction refinement stage that supports document-level inspection and reassignment.
   
Refinement begins from the before--after comparison views and the linked document inspection interface, shown in the supplementary material. Users identify suspicious or ambiguous documents by examining where retrieved documents lie in the projection, inspecting their associated keywords, and, when necessary, opening the full text. Rather than redefining keyword groups, users operate directly on the retrieved document sets produced by the current model update. A document can be reassigned from one concept group to another, or moved into the \texttt{Exclude} group.
These refinement actions provide more targeted supervision than the initial keyword-level specification, especially for documents near concept boundaries. The updated document assignments are then incorporated into the pseudo-supervision set, and KeySI runs another tuning round to adjust the embedding space accordingly.

\section{Usage Scenarios}
We demonstrate \systemname{}'s utility across datasets with different scales and semantic structures: a small COVID-19 corpus (62 documents) and a medium-scale six-class subset of 20 Newsgroups (240 documents). A larger example using the AG News corpus (7,600 documents) can be found in supplemental material.

    \begin{figure*}[tb]
    
    \includegraphics[width=\linewidth,alt={Visualization of the 20 Newsgroups dataset before and after model tuning with KeySI. The left panel shows the user organizing keywords into two concept groups representing space and computers. The center panel shows the initial document embedding projection, where documents from the two concepts overlap and are not clearly separated. The right panel shows the updated embedding projection after tuning, where the two concepts form more distinct clusters. Purple stars indicate the documents associated with the selected concept group, illustrating that the learned concept extends beyond the documents directly used during interaction.}]{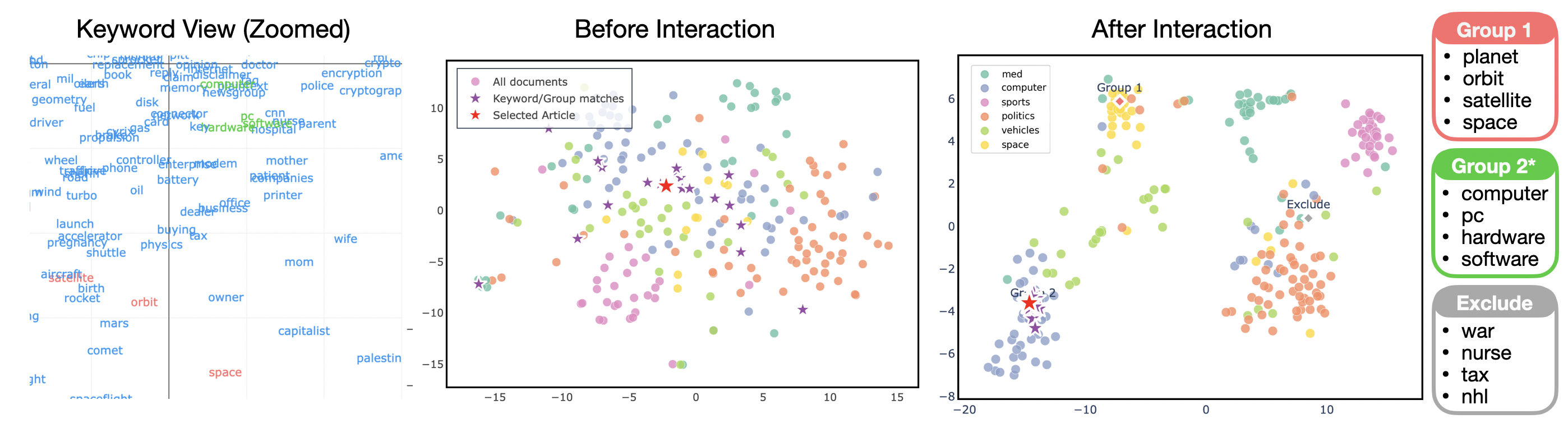}
    \vspace{-2em}
    \caption{\textbf{20 Newsgroups Dataset.} The user specifies two concepts, space and computers, using the keyword view (left), resulting in the groups on the right. In the initial model (center) these concepts are not cleanly structured. After the interaction (right), they are better separated. The purple stars indicate the documents used in the interaction for the selected group (group 2). Notably, in the tuned embeddings, the user-specified concept generalizes well to documents outside of the interaction.}
    \label{fig:newsgroup}
      \vspace{-2em}
\end{figure*}

 \subsection{COVID-19 Risk Factors}
Our first usage scenario is from the perspective of a researcher studying the impacts of smoking and cancer on COVID-19. They have a corpus of 62 research articles on COVID-19 associated with four risk factors: smoking, cancer, neurological conditions, and kidney disease~\cite{COVID19RiskFactor}, and want to create a model that can distinguish between smoking and cancer risk factors for downstream visual analytics. Although these risk factors are semantically distinguishable, the pretrained embedding space exhibits weak structure (Fig.~\ref{fig:teaser}, left). 

The user begins by exploring the keyword view and linked document projection. As they explore, they take note of a group of keywords related to smoking (at the top of Fig.~\ref{fig:UI} (B)) and a group of cancer-related words (at the bottom of Fig.~\ref{fig:UI} (B)). 
 
To assess how well the model captures the concept of smoking, they select the \textit{smoking} keyword and examine its associated documents in the projection using linked interactions. These documents appear dispersed in the document view (Fig.~\ref{fig:UI} (C), purple stars), suggesting that the concept is not yet well captured in the embedding space. They observe a similar pattern when examining cancer-related terms. To provide corrective feedback, they then create two groups (Fig.~\ref{fig:UI} (A)) and begin adding relevant keywords to each group (Fig.~\ref{fig:UI} (B)). Additionally, they add a few keywords from other risk factors to the \texttt{Exclude} group, to keep them separated from their intended concepts. Their final feedback consists of two keyword groups: smoking (\textit{nicotine}, \textit{smoking}, \textit{smoker}) and cancer (\textit{cancer},  \textit{tumour}, \textit{chemotherapy}), and the Exclude group containing \textit{kidney}, \textit{proteinuria}, \textit{seizures} and \textit{epilepsy}, which KeySI then uses to tune the model.

The results of their interaction are shown in Fig.~\ref{fig:teaser}. Note that the ground-truth colors are shown only for demonstration. KeySI presents projections of embeddings before and after tuning, for comparison, with the concept group centroids highlighted to demonstrate their locations. The user sees that the projection now forms two distinct, well-separated groups for the two specified concepts. They also note that the feedback provided enough context to create two additional groups of documents. Using the linked interactions, they verify that these correspond to the other two risk factors included in the \texttt{Exclude} group. Thus, the user was able to efficiently convey their domain knowledge through lightweight keyword interactions that effectively tuned the model.   

\subsection{20 Newsgroups Dataset}
    Our second usage scenario uses a curated six-class subset of the 20 Newsgroups corpus containing 240 documents across six topic groups: medicine, space, politics, sports, computer, and vehicles. This dataset contains topic pairs with uneven levels of semantic separability: some distinctions are relatively clear, whereas others remain overlapping. 

In this example, shown in Fig.~\ref{fig:newsgroup}, the user focuses on a relatively distinguishable pair, space and computers. By exploring the keyword view and linked document projection, the user forms two concept groups using representative keywords for each topic (Fig.~\ref{fig:newsgroup}, right). Upon selecting Group 2 (computers), the user finds that, though the embedding contains some partial topic structure, the retrieved documents are not yet cleanly organized around their intent, demonstrated by the purple stars highlighting the retrieved documents for Group 2 in Fig.~\ref{fig:newsgroup}, left. 

After \systemname{} translates these keyword groups into pseudo-supervision and updates the embedding space, the two target concepts become more clearly separated (Fig.~\ref{fig:newsgroup}, right), although some boundary regions remain mixed. From here, the user can inspect the updated projection and use the refinement interaction to reassign ambiguous documents. An example of this is shown in the supplemental material.

    \section{User Study}
    \label{sec:userstudy}

        We conducted a within-subject user study to examine two aspects of KeySI: (1) whether keyword-based interaction reduces the cost of providing feedback compared to a document-level interaction (DeepSI)~\cite{bian2021deepsi}, and (2) how users form concepts and refine embeddings in an open-ended setting. \add{The study was determined to be exempt by the Tulane University Institutional Review Board (Study 2025-855), and all participants provided informed consent prior to participation.}
    

    \textbf{Baseline.}
    We compare against DeepSI, a document-level interaction in which users drag representative documents into spatial arrangements to convey semantic similarity. 
    We implemented the DeepSI interaction into a basic interface, where users had an interactive projection view to convey feedback and a document view where they could view the contents of selected documents.
     
    
    
    
    \subsection{Participants}
    \label{sec:userstudy_participants}
We recruited $N=8$ participants: four CS PhD students and four participants without CS backgrounds. CS participants reported higher ML familiarity (median $=4.5/7$) and ML usage frequency (median $=4/7$), while non-CS participants reported little or no prior CS/ML experience. Detailed participant demographics and background information are provided in the supplemental material. Although $N=8$ is a small sample, sizes of this range are common in qualitative and mixed-methods visualization studies. The Wilcoxon and binomial results reported in Sec.~5.5.1 therefore provide supporting evidence that the observed differences were pronounced enough to be detectable even at this scale~\cite{caine2016local}.

    \subsection{Study Design and Procedure}
    \label{sec:userstudy_design}
    The study consists of two tasks. In Task A, we used a within-subjects design comparing KeySI and DeepSI on a six-class subset of the 20 Newsgroups dataset~\cite{mitchell1997twentynewsgroups}, counterbalanced for order. Task B was an open-ended exploration using an Amazon Reviews 2023 dataset~\cite{hou2024bridging} (120 documents, four product categories). Sessions lasted one hour and included a tutorial, both tasks with post-task questionnaires, and a final semi-structured interview. 
    



    \subsection{Tasks}
    
    

    \paragraph{Task A: Interaction Comparison.}
Task~A compares keyword-centric and document-centric interaction for text-semantic steering. Our goal is to evaluate the practical cost of concept specification, rather than to compare interfaces or final embedding quality. We hypothesize that keyword-level interaction will lead to lower workload costs than document-centric interactions. Participants were asked to convey two target concepts (e.g., sports and computers) using each interaction.
 In the KeySI condition, participants were instructed to define one keyword group for each concept using the keyword grouping interface. For this task, participants were not provided with the refinement interactions. In the DeepSI condition, participants were instructed to locate representative instances and spatially organize documents based on semantic similarity. During the interactions, participants were asked to explain their reasoning aloud. Based on pilot sessions, we limited participants to 10 minutes for each interaction. We found that this was typically enough time to perform the interaction while keeping the entire session to roughly one hour.


    \paragraph{Task B: Free exploration.}
    Task B evaluated whether keyword-level interaction allows users to achieve clear semantic goals unassisted. Participants were given only the dataset context (Amazon reviews spanning multiple product types) and no labels. They were asked to browse keywords, define concept groups that reflected their semantic intents, trigger training and finally verify and refine results as desired. 
    




    \subsection{Measures}
    
    \textbf{Task A.} For each system, we recorded:
    (i) task completion (binary), whether the participant completed the interaction within 10 minutes
    (ii) time to completion (or time-out at 10 minutes),
    (iii) post-task ratings (7-point Likert) on workload (via NASA-TLX~\cite{hart1988nasa_tlx}), and
    (iv) preference between KeySI and DeepSI with a brief justification.
    \textbf{Task B.} Task~B has no predefined target labels or a single ``correct'' grouping.
    We therefore focused on participants' perceived sensemaking and intent alignment. We logged participants' interactions with KeySI and collected post-task Likert
    ratings on (i) their ability to form a clear semantic goal, (ii) perceived control over the model update, (iii) whether the final
    result reflected their intent, and (iv) refinement usefulness, along with brief open-ended explanations.
    \add{We recorded the distribution of refinement steps. However, due to task heterogeneity, we did not compare reading times as a workload metric.}
    
    \textbf{Interviews.} After completing all tasks, we conducted short semi-structured interviews about system preference, evaluation cues, and perceived limitations.

    \subsection{Task A Results: Interaction Comparison}
    \label{sec:userstudy_taskA_results}

    \subsubsection{Quantitative Results}
    \label{sec:Wil}
    \paragraph{Efficiency} All 8 participants completed the keyword interaction (KeySI) task within the 10-minute limit (mean: 3.35 min), whereas two participants did not fully complete the document interaction (DeepSI) task within
    the same limit (mean: 7.30 min). 
    
    

\begin{figure}[t]
      \centering
      \includegraphics[width=\linewidth,alt={Bar chart comparing subjective workload ratings for KeySI and DeepSI using the six NASA Task Load Index dimensions. Lower ratings indicate lower perceived workload. Bars represent the mean scores with 95 percent confidence intervals. KeySI shows consistently lower workload ratings than DeepSI, with statistically significant differences for the dimensions marked by an asterisk.}]{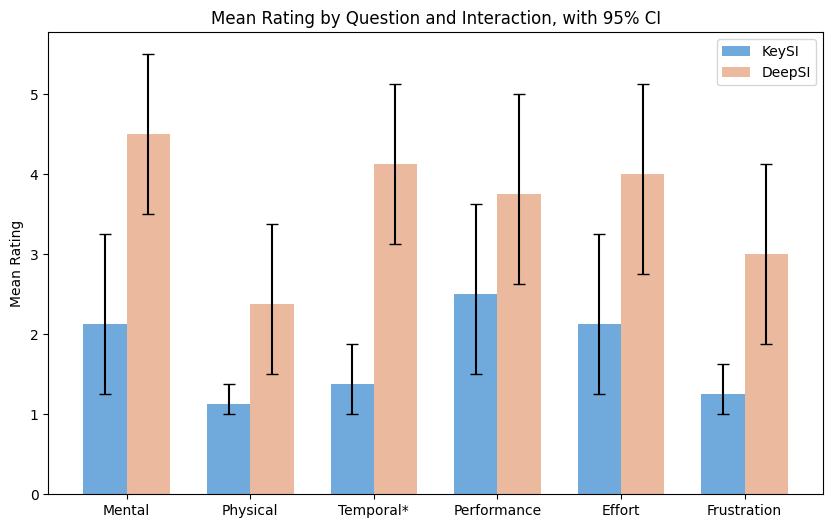}
       \vspace{-1.5em}
      \caption{\textbf{Subjective workload (N=8).}
    Bars show mean ratings for each NASA TLX question (lower is better), with 95\% confidence intervals. \\ * indicates significance after Bonferroni correction.}
      \label{fig:tlx}
       \vspace{-1.5em}
    \end{figure}

\add{  \paragraph{Workload}
We computed an overall NASA-TLX workload score for each participant and condition by averaging the six dimensions, with the performance dimension reverse-coded so that lower values indicate lower workload. \systemname{} produced a lower overall workload score than DeepSI (\systemname{}: M=1.75; DeepSI: M=3.63; Wilcoxon signed-rank test: W=0, p=.004; Fig.~\ref{fig:tlx}).
At the dimension level, \systemname{} showed lower ratings for mental demand (W=4, p=.031), physical demand (W=0, p=.031), temporal demand (W=0, p=.004), effort (W=0, p=.031), and frustration (W=0, p=.016), and higher perceived performance (W=0, p=.031). After Bonferroni correction across the six dimension-level tests, temporal demand remained significant ($p_{\mathrm{corr}}=.024$). Together, these results support our hypothesis that keyword-level interactions have lower costs to the user than document-level interactions.}

    \paragraph{Preference}
    Participants preferred keyword interactions over document interactions for overall preference (8/8), intuitiveness (8/8), and real-world usefulness (8/8), with each significant by exact binomial test (p=.004). The only item where DeepSI was selected more often, though not significantly (5/8 DeepSI vs.\ 3/8 KeySI; p=.727), was helping users understand the topics, suggesting that direct access to full documents sometimes provided richer context.
    
    \add{Because the CS and non-CS subgroups each contained only four participants, we treat background differences as exploratory. Preferences were broadly consistent across backgrounds. For example, both groups reported high agreement that KeySI helped them form semantic goals (CS: M=6.0, non-CS: M=6.25) and understand the structure of the data (CS: M=6.25, non-CS: M=6.75). Participants from both backgrounds also rated the refinement view positively (CS: M=6.75, non-CS: M=6.50). These observations suggest similar trends across backgrounds, though larger studies are needed.}
      

    


\subsubsection{Qualitative Feedback}
\label{sec:taska_qual}
 Participants generally described \systemname{} as lower-effort and more intuitive because it allowed them to express a concept directly through keywords, rather than first locating representative documents. As one participant noted, the keyword workflow was ``very intuitive'' because they could ``just put the keywords I want into groups'' (P3). Participants also reported that keyword-based interaction supported rapid judgments without opening many documents (P1, P3-P8). Together with the perceived workload reported above, these comments suggest that \systemname{} reduces the cost of translating a semantic goal into usable supervision, which helps explain its lower perceived workload in the study task. 

In contrast, participants attributed DeepSI's higher workload to the need to inspect documents before meaningful interaction (P1-P4, P6). As one participant stated, ``the biggest challenge is I need to read the document and think about it. That needs a lot of time'' (P3). At the same time, this document-centric workflow also offered an important advantage: full documents provided richer context for understanding topics (P2, P5, P6-P8) and the ability to directly handle ambiguous cases (P6, P7, P8). However, participants also described using fast reading strategies that relied on scanning for salient terms rather than fully reading each document (P6, P7). This suggests that even within a document-centric workflow, users often rely on keyword-like cues to judge relevance. In this sense, \systemname{} makes this implicit strategy explicit by allowing users to express concepts directly through keywords. This helps explain the pattern above: \systemname{} was generally preferred overall, while DeepSI was selected more often for topic understanding. Taken together, these results suggest a trade-off between lower-cost concept specification and richer document-level context.

Participants also noted both limitations and strengths of \systemname{}. One participant pointed out that noise in the keyword pool can make it harder to locate useful terms: “If the noisy keywords were filtered out in advance, it would be much easier to find and select the useful ones for defining my concept” (P2). The participants also indicated that more specific words (P1, P3) and more structure in the visual layout (P2, P5), such as categories, would make it easier to find keywords.  In contrast, another participant highlighted the scalability advantage of keyword-level interaction over document-by-document inspection: “I prefer the keyword-based workflow. When the number of documents grows, reading them one by one is almost impossible, but with keywords I can quickly select many documents that share a topic” (P3). Together, these comments suggest that keyword-level interaction can lower the entry cost of model steering at scale, while the usability of the keyword view still depends on effective keyword curation.

\add{We also observed quantitative improvements in the high-dimensional in-session tuned embeddings, even when participants selected different keywords for the same target concept pair. Specifically, retrieved documents associated with the same target concept became more coherent according to neighborhood-based purity measures computed in the original embedding space. This suggests that \systemname{} is not tied to one fixed keyword formulation, but can produce similar concept-alignment effects across different keyword choices for the same semantic goal. Detailed results are reported in the supplemental material.}
 

    \subsection{Task B Results}
    \label{sec:userstudy_taskb_results}
    
    
    \subsubsection{Perceived Sensemaking and Intent Alignment}
    Participants provided Likert feedback (on a 7-point scale) describing their perceived ability to use KeySI. 
    Participants generally reported that they could (i) form a clear semantic goal (median=6.0), (ii) use the keyword-based interaction to explore and understand the dataset structure (median=7.0), and (iii) influence how the model organized documents (median=6.5). Participants tended to agree that the final result reflected their feedback and intentions (median=6.0). Overall, these responses provide preliminary evidence that KeySI supports intent expression and verification in a realistic exploratory workflow where user concepts are not necessarily aligned with the data's original labels.




\subsubsection{Qualitative explanations}
\label{sec:taskb_qual}

Task~B reflects open-ended semantic steering without predefined target labels or a
single ``correct'' grouping. Our findings are grounded in session recordings and
interaction logs, together with participants' think-aloud and interview remarks.

Overall, we observed that participants did not specify a complete concept definition upfront.
Instead, they converged iteratively: they began with a rough semantic direction,
inspected retrieved candidates and keyword previews, and then adjusted keywords
or group membership as their intent became clearer. For example, one participant
reported revising their keyword choice after noticing that the retrieved results
``still looked mixed'' and did not match what they intended (P2).

    \begin{table}[t]
      \centering
      \setlength{\tabcolsep}{6pt}
      \begin{tabular}{l|ccc|cc}
        \toprule
        \multirow{2}{*}{\textbf{Pair}}& \textbf{Purity} & \textbf{Purity} & \multirow{2}{*}{\textbf{Purity $\Delta$ }} &  \multirow{2}{*}{\textbf{NRR}} &  \multirow{2}{*}{\textbf{TPL}} \\
        &(lexical) & (denoised)  \\
        \midrule
        P1 & 0.43 & 0.61 & 0.18 &  0.56 & 0.17\\
        P2 & 0.63 & 0.75 & 0.12 & 0.43 & 0.06 \\
        P3   & 0.61 & 0.83 & 0.21 & 0.64 & 0.14\\
        P4   & 0.71 & 0.92 & 0.21 & 0.83 & 0.14\\
        P5    & 0.75 & 0.93 & 0.18 & 0.44 & 0.10\\
        \bottomrule
      \end{tabular}
       \vspace{-0.5em}
      \caption{\textbf{Effects of Semantic Denoising.} Results are shown for five keyword-group pairs (P1–P5) under the stress-test setting described in Sec.~\ref{sec:expr2}. The first two columns report the baseline purity using lexical retrieval and the purity after gap-based semantic denoising (computed via ground-truth labels), while the third reports the improvement. The last two columns report the proportion of noise (NRR) and true positives (TPL) discarded from the candidate document set during semantic denoising.}
      \label{tab:denoising}
      \vspace{-1.5em}
    \end{table}

Refinement was commonly used to correct scope mismatches that emerged during
exploration. In one case, a participant initially used \emph{bag} to form a group
for accessories, but then noticed that documents about dog bags were
also retrieved; they moved those documents into \texttt{Exclude} to better match
their intended scope (P6, P8). Participants liked that the refinement view ``allowed me to solidify the subjects that I'm looking for'' (P6). 
This behavior aligns with the most commonly reported refinement goals of reducing mixed documents and clarifying boundaries.
Participants most often wanted to reduce mixed documents (6/8), sharpen
boundaries (6/8), and improve within-group consistency (6/8), consistent with
their self-described use of refinement as a boundary-cleaning step. In addition to self-reported remarks, our logs revealed a
consistent refinement strategy: participants often clicked visually isolated
points (far from dense regions) in the projection, checked the keyword preview (opening the full text as needed) to judge relevance, reassigned the item
to a better-matching group, and re-ran training. This suggests that the
projection structure itself serves as an important cue for locating ``samples to
correct'' during open-ended exploration.
Participants generally stopped refining when the result became sufficiently interpretable and adequately matched their intent, rather than when it reached a single objectively ``correct'' grouping. This suggests that, in an open-ended setting, refinement functions primarily as a sensemaking and boundary-cleaning process.

 \section{Quantitative Evaluation}
    \label{sec:quant}
    This section conducts two quantitative experiments to evaluate the effectiveness of KeySI at producing strong supervision signals and restructuring the embedding space. The experiments evaluate the following two questions: \textbf{RQ 1:} Does semantic denoising create stronger supervision signals, and \textbf{RQ 2:} Compared with the initial embeddings, how well does the pseudo-supervision globally re-structure the embedding space to reflect user feedback (i.e., beyond just the retrieved documents involved in the interaction)? 
    
    
    

 \begin{figure}[t]
      \centering
      \includegraphics[width=\linewidth,alt={Line chart showing concept-set purity at k equals 10 before and after tuning for multiple dataset and task pairs. Each line connects the pre-tuning and post-tuning purity values for one dataset or task. Most lines slope upward, indicating that concept-set purity improves after tuning. Higher values correspond to better concept purity.}]{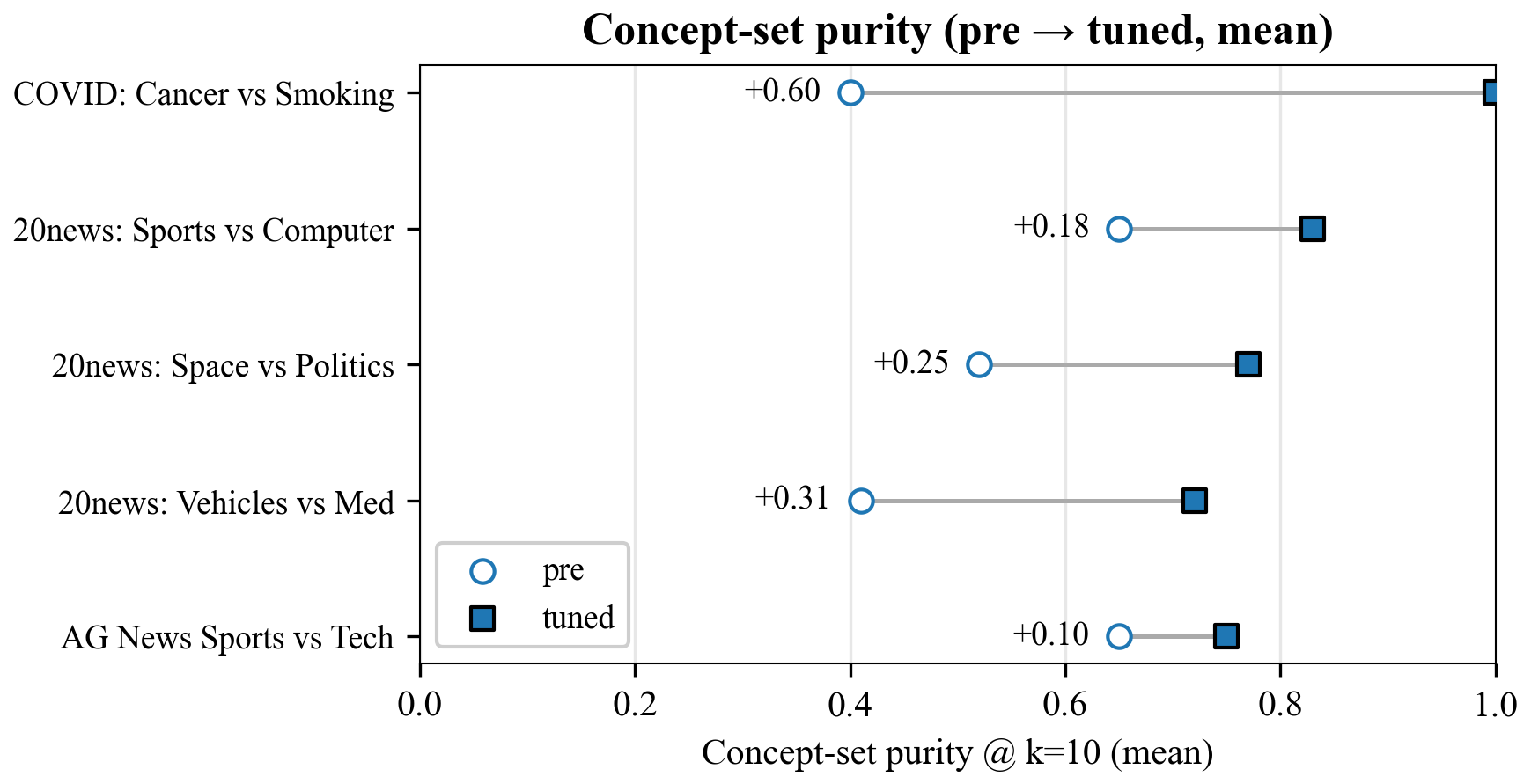}
      \vspace{-1em}
      \caption{\textbf{Concept-set purity at $k{=}10$ (pre $\rightarrow$ tuned).}
Each line shows one dataset/task pair. Higher is better.}
      \label{fig:concept_purity_at10}
      \vspace{-2em}
    \end{figure}

    \subsection{Experiment 1: Supervision Signal Quality}
    This experiment addresses RQ1 by evaluating the quality of the supervision signals retrieved from the keywords before and after applying semantic denoising, in terms of how many on-topic (true positive) vs how many noisy (false positive) documents it retrieves. The goal is to quantify the impact of semantic denoising and demonstrate that it produces more relevant signals.  
    
    \textbf{Setup:} To evaluate whether semantic denoising removes noisy matches, we use the same curated six-class subset of 20 Newsgroups as in the usage scenario. We report results on five keyword-group pairs. Each keyword-group corresponds to a label in the dataset, which we use for evaluation. For each pair, we construct two concept groups and one \texttt{Exclude} group from the corpus keyword pool. To simulate imperfect keyword selection, we inject one off-topic keyword into each concept group. If successful, semantic denoising should focus on the semantically coherent keywords (those corresponding to the ground-truth label) and filter noisy matches introduced by the off-topic keyword. 

    \textbf{Metrics:} We report three metrics: Purity (true-positive proportion within the pseudo-supervision set; higher is better), Noise Removal Rate (NRR; proportion of false positives removed by semantic denoising; higher is better), and True-Positive Loss (TPL; proportion of true positives removed by semantic denoising; lower is better).

    \textbf{Results:}  Table~\ref{tab:denoising} shows that semantic denoising consistently improves pseudo-supervision purity across all five keyword-group pairs, with gains of about 20 percentage points. It further shows why this improvement occurs: semantic denoising removes a substantial portion of noisy candidates across all pairs, while discarding only a relatively small portion of true positives. Together, these results show that semantic denoising produces cleaner supervision for tuning.
    
    This matters because the filtered document sets are used to update the model. Cleaner supervision gives a more reliable training signal, while noisy documents can weaken the intended concept boundaries. At the same time, semantic denoising is not perfect: when keyword groups are too broad or noisy, some relevant documents may also be removed. In such cases, KeySI allows users to inspect and correct the result through the interaction refinement. We repeated each experiment five times with different random seeds and report the mean.

    \subsection{Experiment 2: Embedding Structure}
    \label{sec:expr2}
    This experiment addresses RQ2 by evaluating the quality of the tuned embeddings, relative to the ground truth labels. The goal is to demonstrate that the model update does not simply memorize the documents retrieved by KeySI, but captures the intent and applies it globally. A successful result creates an embedding space where documents whose ground-truth matches the target concept become grouped together. 
    
    \textbf{Setup:} To evaluate whether tuning improves embedding structure under different keyword-group specifications, we use the same datasets as in the usage scenarios: COVID-19, the curated 20 Newsgroups subset, and AG News. For each dataset, we run the full \systemname{} pipeline and compare the embeddings before and after tuning. \add{Runtime statistics for these pipeline runs are reported in the supplemental material.}

 \textbf{Metrics:} To measure whether neighborhoods around the target concepts become cleaner after tuning, we compute the purity of the $k$ nearest neighbors, with $k=10$, for a queried point by measuring the fraction of documents of the same ground truth label among its 10 nearest neighbors. 
    We define concept-set purity as the $k$NN purity using all documents in the concept's ground truth as queries, as opposed to user-set purity that only uses the subset directly selected or refined by the user. We also 
    report a silhouette score\cite{rousseeuw1987silhouettes} computed on the full two-topic 
    ground-truth sets to quantify their separability in the embedding space. \add{Both metrics are computed in the original high-dimensional embedding space. }
    Since the majority of these documents were never directly supervised 
    during tuning, improvement in these metrics reflects generalization of 
    the learned embedding structure rather than memorization of the training signal. 
    For brevity, we report concept-set purity at $k{=}10$ here; user-set purity, results for other $k$ values, and full per-pair breakdowns are provided in the supplemental material.

    \textbf{Results:} Figure~\ref{fig:concept_purity_at10} shows that KeySI tuning improves neighborhood coherence for the two target concepts across all evaluated dataset/task pairs. Using $k{=}10$ nearest neighbors, concept-set purity increases consistently after tuning, indicating that the improvement extends beyond the user-provided supervision set to other documents belonging to the same target concepts. This suggests that the updated embedding space better captures the broader semantic structure of the intended concepts, rather than improving only the directly supervised instances. 

     Table~\ref{tab:silhouette} reports silhouette scores computed on the two target concepts. The increase in scores across all pairs indicates a clearer separation between the two target concepts in the tuned embeddings. Together, these results provide quantitative evidence that KeySI tuning improves both concept-level neighborhood coherence and between-concept separability. Importantly, these improvements are obtained from a small number of keyword-level interaction steps, indicating that even low-cost interaction can produce signals strong enough to meaningfully restructure the embedding space. This suggests that KeySI achieves a favorable trade-off between interaction cost and embedding effectiveness.

    \section{Discussion}
    \label{sec:discussion}

\begin{table}[t]
  \centering
  \setlength{\tabcolsep}{6pt}
  \begin{tabular}{lcc}
    \toprule
    \textbf{Pair} & \textbf{Pre (mean)} & \textbf{Tuned (mean)} \\
    \midrule
    20news: Sports vs Computer & 0.18 & 0.62 \\
    20news: Space vs Politics  & 0.05 & 0.59 \\
    20news: Vehicles vs Med    & 0.08 & 0.48 \\
    COVID: Cancer vs Smoking   & 0.06 & 0.97 \\
    AG: Sports vs Sci/Tech     & 0.09 & 0.79 \\
    \bottomrule
  \end{tabular}
  \vspace{-0.5em}
  \caption{\textbf{Silhouette scores (pre vs.\ tuned; mean).} Higher is better.
  Values are averaged across repeated runs for each pair.}
  \label{tab:silhouette}
  \vspace{-2em}
\end{table}

    \paragraph{Keyword-level vs Document-level Interaction}
    Document-centric interactions require users to first locate representative documents and then use those instances to indirectly communicate a higher-level concept. Our study suggests that this reverses the way many users reason: they often begin with a concept, and then look for documents that fit it. This mismatch results in slower specification and higher effort in document-level interaction. In contrast, KeySI shifts the interaction paradigm from document manipulation to keyword-level concept specification, allowing users to express intent more directly through concept-relevant terms. However, participants in our study still valued the ability to view and manipulate documents directly, suggesting that users might benefit from a hybrid interaction that operates in both document and keyword spaces, balancing rapid keyword-based selections with direct manipulations of retrieved documents. Our refinement touches this, but future research is needed to investigate it directly. Our comparison used the original DeepSI interface without retrieval assistance. Therefore, the observed differences may not generalize to document-level interaction systems with stronger search capabilities.
    
    \paragraph{Keyword Selection Considerations}\add{Our study found that keyword usefulness depends on the user's concept distinction. Overly general keywords can reduce concept quality when they describe multiple target concepts. For example, if users aim to distinguish cancer-related documents from neurological-disease documents, a broad term such as ``disease'' is not sufficiently discriminative because it may apply to both groups. Users should select keywords that match the desired semantic distinction and use the \texttt{Exclude} group to remove out-of-scope directions. The supplemental material provides further guidance.

The current keyword-selection process relies on KeyBERT-based extraction and frequency-based curation. As corpus size increases, the number of keywords may grow substantially, making users likely to be overwhelmed by domain-generic or low-utility terms and increasing the effort needed to identify useful concept keywords. Raising the frequency threshold can reduce the keyword pool, but it may also remove rare yet discriminative terms that are important for specific concepts. This creates a trade-off between reducing visual search burden and preserving concept-specific keywords. Although the supplemental material demonstrates the pipeline on a 7,600-document corpus, we didn't conduct a formal user study at this scale. Future work should investigate adaptive keyword ranking and filtering strategies that balance frequency, specificity, and user intent. Additionally, it should evaluate the usability of KeySI for a large corpus, particularly the effects of larger keyword pools and document matches on user interaction.}


    \paragraph{Feedback Inspectability}
    Based on our user study, interactive model steering requires not only feedback input, but also feedback inspectability. Model updates are abstract and not directly observable, so users must be able to judge whether the system has interpreted their feedback appropriately. Our results suggest that inspectability is essential rather than optional. Participants did not simply provide keyword groups and accept the result; they relied on retrieved candidates, keyword summaries, before--after projections, and refinement to evaluate whether the update matched their intent. This suggests that effective human-in-the-loop model steering should not be a one-shot interaction, but an inspectable and correctable workflow in which users can both provide feedback and verify how it is incorporated into the model. Our prototype provides light-weight interactions to assist in inspection tasks, but future work is needed to provide richer context and deeper connections between both the document and text feature spaces and the pre/post tuning embeddings to better guide interactions. 

\paragraph{Generalizability of Interaction Pipeline}
    Our implementation adopts BERT for document embeddings, uses KeyBERT for keyword extraction, and applies a gap-based semantic denoising strategy to filter out semantically irrelevant documents matched based on keyword groups. \add{We use t-SNE as the default projection because it emphasizes local neighborhood preservation, which supports visual inspection of concept coherence. However, regardless of the dimensionality reduction method, users still interact with a 2D projection of high-dimensional embeddings, so their decisions may be influenced by projection artifacts. Future work should examine how projection uncertainty affects user interpretation and steering decisions.} Considering the diversity of data characteristics, tasks, and user preferences, we deliberately designed a flexible interaction paradigm that allows users to customize the methods for keyword extraction, dimensionality reduction, and embedding according to their specific needs. 

 \paragraph{Interactions \& Loss Functions}
    \systemname{} opts for a cluster-focused interaction, through the definition of disjoint keyword groups and the application of Triplet Margin Loss, a cluster-focused loss function. Though this provides an intuitive means for conveying feedback, it may not encompass all user tasks. For instance, it does not support defining inter-topic relationships beyond very discrete clustering. However, some tasks may be better supported through more complex relationships - e.g., through the definition of a hierarchy of topics or explicit cluster similarities. Future work should explore designing intuitive interactions and corresponding loss functions for more complex feedback.
   
\add{\paragraph{Downstream Task Impact}KeySI adapts the embedding space toward user-specified semantic distinctions rather than general-purpose task performance. While our evaluation shows improvements for the specified concepts, future work should examine how tuning affects downstream tasks such as retrieval, classification, or clustering.}

    \section{Conclusion}
    \label{sec:conclusion}
We presented \systemname{}, a keyword-centric semantic interaction framework for text that treats concept specification as the primary interaction for model steering, rather than document manipulation. Across a user study, quantitative validation, and usage scenarios, our findings suggest that keyword-level interaction provides a more direct and lower-cost way to express semantic intent in early-stage text analysis. The translation pipeline and refinement workflow support this interaction model by making concept-level input learnable, inspectable, and correctable.

\acknowledgments{

Generative AI tools were used for language editing only. All substantive research decisions, analyses, interpretations, and writing were conducted by the authors. We thank the participants in our user study for their time and valuable feedback.

}
\section*{Supplementary Material}

The project repository, including source code, corpus, and a demonstration video, is available at:
\url{https://github.com/Tulane-Vis-Research/KeySI}

The supplemental document includes additional implementation details, experimental results, participant information, questionnaire materials, statistical analyses, and extended visualizations that complement the main paper.

\bibliographystyle{abbrv-doi-hyperref}

\bibliography{template}

@inproceedings{bekkerman2007interactive,
  author    = {Bekkerman, Ron and Raghavan, Hema and Allan, James and Eguchi, Koji},
  booktitle = {IJCAI},
  pages     = {684--689},
  title     = {Interactive Clustering of Text Collections According to a User-Specified Criterion},
  volume    = {7},
  year      = {2007}
}

@inproceedings{bian2019deepva,
  author       = {Bian, Yali and Wenskovitch, John and North, Chris},
  booktitle    = {2019 IEEE Workshop on Machine Learning from User Interaction for Visualization and Analytics (MLUI)},
  doi          = {10.1109/mlui52769.2019.10075565},
  organization = {IEEE},
  pages        = {1--10},
  title        = {Deepva: Bridging Cognition and Computation Through Semantic Interaction and Deep Learning},
  year         = {2019}
}

@inproceedings{bian2021deepsi,
  author    = {Bian, Yali and North, Chris},
  booktitle = {Proc. 26th International Conference on Intelligent User Interfaces},
  doi       = {10.1145/3397481.3450670},
  pages     = {197--207},
  title     = {Deepsi: Interactive Deep Learning for Semantic Interaction},
  year      = {2021}
}

@article{bian2024neuralsi,
  author  = {Bian, Yali and Faust, Rebecca and North, Chris},
doi = {10.48550/arXiv.2402.17178},
  journal = {arXiv preprint arXiv:2402.17178},
  title   = {NeuralSI: Neural Design of Semantic Interaction for Interactive Deep Learning},
  year    = {2024}
}

@inproceedings{bird2006nltk,
  author    = {Bird, Steven},
  booktitle = {Proc. COLING/ACL interactive presentation sessions},
  doi       = {10.3115/1225403.1225421},
  pages     = {69--72},
  title     = {NLTK: the natural language toolkit},
  year      = {2006}
}

@inproceedings{bradel2015big,
  author       = {Bradel, Lauren and Wycoff, Nathan and House, Leanna and North, Chris},
  booktitle    = {2015 Big Data Visual Analytics (BDVA)},
  doi          = {10.1109/bdva.2015.7314287},
  organization = {IEEE},
  pages        = {1--8},
  title        = {Big text visual analytics in sensemaking},
  year         = {2015}
}

@inproceedings{caine2016local,
  address   = {New York, NY},
  author    = {Caine, Kelly},
  booktitle = {Proc. {SIGCHI} Conference on Human Factors in Computing Systems},
  doi       = {10.1145/2858036.2858498},
  pages     = {981--992},
  publisher = {ACM},
  title     = {Local Standards for Sample Size at {CHI}},
  year      = {2016}
}

@misc{COVID19RiskFactor,
  author   = {Allen Institute For AI and Anthony Goldbloom and Ben Hamner and Carissa Schoenick and Timo Bozsolik and Paul Mooney and Peijen Lin and Sebastian Kohlmeier and devrishi},
  language = {en},
  title    = {{COVID-19 Open Research Dataset Challenge (CORD-19)}},
  url      = {https://www.kaggle.com/datasets/allen-institute-for-ai/CORD-19-research-challenge},
  urldate  = {2025-03-31},
  year     = {2022}
}

@article{dowling2019interactive,
  author    = {Dowling, Michelle and Wycoff, Nathan and Mayer, Brian and Wenskovitch, John and House, Leanna and Polys, Nicholas and North, Chris and Hauck, Peter},
  doi       = {10.1016/j.bdr.2019.04.003},
  journal   = {Big Data Research},
  pages     = {49--58},
  publisher = {Elsevier},
  title     = {Interactive visual analytics for sensemaking with big text},
  volume    = {16},
  year      = {2019}
}

@inproceedings{endert2012semantic,
  author    = {Endert, Alex and Fiaux, Patrick and North, Chris},
  booktitle = {Proc. SIGCHI conference on Human factors in computing systems},
  doi       = {10.1145/2207676.2207741},
  pages     = {473--482},
  title     = {Semantic interaction for visual text analytics},
  year      = {2012}
}

@article{gorg2012combining,
  author    = {G{\"o}rg, Carsten and Liu, Zhicheng and Kihm, Jaeyeon and Choo, Jaegul and Park, Haesun and Stasko, John},
  doi       = {10.1109/tvcg.2012.324},
  journal   = {IEEE transactions on Visualization and Computer Graphics},
  number    = {10},
  pages     = {1646--1663},
  publisher = {IEEE},
  title     = {Combining computational analyses and interactive visualization for document exploration and sensemaking in jigsaw},
  volume    = {19},
  year      = {2012}
}

@misc{grootendorst2020keybert,
  author    = {Maarten Grootendorst},
  doi       = {10.5281/zenodo.4461265},
  publisher = {Zenodo},
  title     = {KeyBERT: Minimal keyword extraction with BERT},
  version   = {v0.3.0},
  year      = {2020}
}

@incollection{hart1988nasa_tlx,
  author    = {Hart, Sandra G. and Staveland, Lowell E.},
  booktitle = {Human Mental Workload},
  doi       = {10.1016/S0166-4115(08)62386-9},
  editor    = {Hancock, Peter A. and Meshkati, Najmedin},
  pages     = {139--183},
  publisher = {North-Holland},
  title     = {Development of NASA-TLX (Task Load Index): Results of Empirical and Theoretical Research},
  year      = {1988}
}

@misc{ho2024surveypretrainedlanguagemodels,
  archiveprefix = {arXiv},
  author        = {Xanh Ho and Anh Khoa Duong Nguyen and An Tuan Dao and Junfeng Jiang and Yuki Chida and Kaito Sugimoto and Huy Quoc To and Florian Boudin and Akiko Aizawa},
  doi           = {10.48550/arXiv.2401.17824},
  eprint        = {2401.17824},
  primaryclass  = {cs.CL},
  title         = {A Survey of Pre-Trained Language Models for Processing Scientific Text},
  url           = {https://arxiv.org/abs/2401.17824},
  year          = {2024}
}

@article{hou2024bridging,
  author  = {Hou, Yupeng and Li, Jiacheng and He, Zhankui and Yan, An and Chen, Xiusi and McAuley, Julian},
  doi     = {10.48550/arXiv.2403.03952},
  journal = {arXiv preprint arXiv.2403.03952},
  title   = {Bridging Language and Items for Retrieval and Recommendation},
  year    = {2024}
}

@inproceedings{hu2011interactive,
  author    = {Hu, Yeming and Milios, Evangelos E and Blustein, James},
  booktitle = {Proc. ACM symposium on applied computing},
  doi       = {10.1145/1982185.1982436},
  pages     = {1143--1150},
  title     = {Interactive feature selection for document clustering},
  year      = {2011}
}

@article{hu2014interactive,
  author    = {Hu, Yeming and Milios, Evangelos E and Blustein, James},
  doi       = {10.3233/ida-140658},
  journal   = {Intelligent Data Analysis},
  number    = {4},
  pages     = {561--581},
  publisher = {SAGE Publications Sage UK: London, England},
  title     = {Interactive document clustering with feature supervision through reweighting},
  volume    = {18},
  year      = {2014}
}

@article{kang2012examining,
  author    = {Kang, Youn-ah and Stasko, John},
  doi       = {10.1109/tvcg.2012.224},
  journal   = {IEEE Transactions on Visualization and Computer Graphics},
  number    = {12},
  pages     = {2869--2878},
  publisher = {IEEE},
  title     = {Examining the Use of a Visual Analytics System for Sensemaking Tasks: Case Studies with Domain Experts},
  volume    = {18},
  year      = {2012}
}

@article{kiefer2022semantic,
  author    = {Kiefer, Sebastian and Hoffmann, Mareike and Schmid, Ute},
  doi       = {10.3390/make4040050},
  journal   = {Machine Learning and Knowledge Extraction},
  number    = {4},
  pages     = {994--1010},
  publisher = {MDPI},
  title     = {Semantic interactive learning for text classification: a constructive approach for contextual interactions},
  volume    = {4},
  year      = {2022}
}

@inproceedings{lam2024concept,
  author    = {Lam, Michelle S. and Teoh, Janice and Landay, James A. and Heer, Jeffrey and Bernstein, Michael S.},
  booktitle = {Proc. CHI Conference on Human Factors in Computing Systems},
  doi       = {10.1145/3613904.3642830},
  publisher = {ACM},
  title     = {Concept Induction: Analyzing Unstructured Text with High-Level Concepts Using LLooM},
  year      = {2024}
}

@inproceedings{lee2012ivisclustering,
  author       = {Lee, Hanseung and Kihm, Jaeyeon and Choo, Jaegul and Stasko, John and Park, Haesun},
  booktitle    = {Computer graphics forum},
  doi          = {10.1111/j.1467-8659.2012.03108.x},
  number       = {3pt3},
  organization = {Wiley Online Library},
  pages        = {1155--1164},
  title        = {iVisClustering: An interactive visual document clustering via topic modeling},
  volume       = {31},
  year         = {2012}
}

@inproceedings{lin2024imagesi,
  author       = {Lin, Jiayue and Faust, Rebecca and North, Chris},
  booktitle    = {IEEE Visualization and Visual Analytics (VIS)},
  doi          = {10.1109/vis55277.2024.00026},
  organization = {IEEE},
  pages        = {91--95},
  title        = {ImageSI: Semantic Interaction for Deep Learning Image Projections},
  year         = {2024}
}

@inproceedings{liu2009interactive,
  author    = {Liu, Shixia and Zhou, Michelle X and Pan, Shimei and Qian, Weihong and Cai, Weijia and Lian, Xiaoxiao},
  booktitle = {Proc. 18th ACM conference on Information and knowledge management},
  doi       = {10.1145/1645953.1646023},
  pages     = {543--552},
  title     = {Interactive, topic-based visual text summarization and analysis},
  year      = {2009}
}

@article{maaten2008visualizing,
  author  = {van der Maaten, Laurens and Hinton, Geoffrey},
  journal = {Journal of Machine Learning Research},
  number  = {86},
  pages   = {2579--2605},
  title   = {Visualizing Data using t-SNE},
  volume  = {9},
  year    = {2008}
}

@inproceedings{mishra2021designing,
  author    = {Mishra, Swati and Rzeszotarski, Jeffrey M},
  booktitle = {Proc. CHI Conference on Human Factors in Computing Systems},
  doi       = {10.1145/3411764.3445096},
  pages     = {1--15},
  title     = {Designing Interactive Transfer Learning Tools for ML Non-Experts},
  year      = {2021}
}

@misc{mitchell1997twentynewsgroups,
  author    = {Tom Mitchell},
  publisher = {UCI Machine Learning Repository},
  title     = {Twenty Newsgroups},
  url       = {https://archive.ics.uci.edu/dataset/113/twenty+newsgroups},
  year      = {1997}
}

@inproceedings{nourashrafeddin2013interactive,
  author    = {Nourashrafeddin, Seyednaser and Milios, Evangelos and Arnold, Dirk},
  booktitle = {Proc. ACM symposium on Document Engineering},
  doi       = {10.1145/2494266.2494279},
  pages     = {61--70},
  title     = {Interactive text document clustering using feature labeling},
  year      = {2013}
}

@inproceedings{rezaeipourfarsangi2022interactive,
  author    = {Rezaeipourfarsangi, Sima and Pei, Ningyuan and Sherkat, Ehsan and Milios, Evangelos},
  booktitle = {Proc. International Conference on Advanced Visual Interfaces},
  doi       = {10.1145/3531073.3531174},
  pages     = {1--5},
  title     = {Interactive clustering and high-recall information retrieval using language models},
  year      = {2022}
}

@article{rousseeuw1987silhouettes,
  author    = {Rousseeuw, Peter J},
  doi       = {10.1016/0377-0427(87)90125-7},
  journal   = {Journal of computational and applied mathematics},
  pages     = {53--65},
  publisher = {Elsevier},
  title     = {Silhouettes: a graphical aid to the interpretation and validation of cluster analysis},
  volume    = {20},
  year      = {1987}
}

@article{ruppert2017visual,
  author    = {Ruppert, Tobias and Staab, Michael and Bannach, Andreas and L{\"u}cke-Tieke, Hendrik and Bernard, J{\"u}rgen and Kuijper, Arjan and Kohlhammer, J{\"o}rn},
  doi       = {10.2352/issn.2470-1173.2017.1.vda-388},
  journal   = {Electronic Imaging},
  pages     = {46--57},
  publisher = {Society for Imaging Science and Technology},
  title     = {Visual interactive creation and validation of text clustering workflows to explore document collections},
  volume    = {29},
  year      = {2017}
}

@inproceedings{sakai2004multiple,
  author    = {Sakai, Hiroyuki and Masuyama, Shigeru},
  booktitle = {Proc. 20th International Conference on Computational Linguistics},
  doi       = {10.3115/1220355.1220499},
  pages     = {1001--1007},
  title     = {A multiple-document summarization system with user interaction},
  year      = {2004}
}

@inproceedings{sen2019heidl,
  author    = {Sen, Prithviraj and Li, Yunyao and Kandogan, Eser and Yang, Yiwei and Lasecki, Walter},
  booktitle = {Proc. 57th Annual Meeting of the Association for Computational Linguistics: System Demonstrations},
  doi       = {10.18653/v1/p19-3023},
  pages     = {135--140},
  title     = {HEIDL: Learning Linguistic Expressions with Deep Learning and Human-in-the-Loop},
  year      = {2019}
}

@inproceedings{sherkat2018interactive,
  author    = {Sherkat, Ehsan and Nourashrafeddin, Seyednaser and Milios, Evangelos E and Minghim, Rosane},
  booktitle = {Proc. 23rd International Conference on Intelligent User Interfaces},
  doi       = {10.1145/3172944.3172964},
  pages     = {281--292},
  title     = {Interactive Document Clustering Revisited: A Visual Analytics Approach},
  year      = {2018}
}

@article{sherkat2019visual,
  author    = {Sherkat, Ehsan and Milios, Evangelos E and Minghim, Rosane},
  doi       = {10.1145/3241380},
  journal   = {ACM Transactions on Interactive Intelligent Systems (TiiS)},
  number    = {1},
  pages     = {1--33},
  publisher = {ACM New York, NY, USA},
  title     = {A visual analytics approach for interactive document clustering},
  volume    = {10},
  year      = {2019}
}

@inproceedings{stasko2007jigsaw,
  author       = {Stasko, John and Gorg, Carsten and Liu, Zhicheng and Singhal, Kanupriya},
  booktitle    = {IEEE Symposium on Visual Analytics Science and Technology},
  doi          = {10.1109/vast.2007.4389006},
  organization = {IEEE},
  pages        = {131--138},
  title        = {Jigsaw: supporting investigative analysis through interactive visualization},
  year         = {2007}
}

@article{wang2023pre,
  author    = {Wang, Haifeng and Li, Jiwei and Wu, Hua and Hovy, Eduard and Sun, Yu},
  doi       = {10.1016/j.eng.2022.04.024},
  journal   = {Engineering},
  pages     = {51--65},
  publisher = {Elsevier},
  title     = {Pre-trained language models and their applications},
  volume    = {25},
  year      = {2023}
}

@inproceedings{yang2019study,
  author    = {Yang, Yiwei and Kandogan, Eser and Li, Yunyao and Sen, Prithviraj and Lasecki, Walter S},
  booktitle = {IUI Workshops},
  title     = {A Study on Interaction in Human-in-the-Loop Machine Learning for Text Analytics},
  year      = {2019}
}

@article{choo2013utopian,
  author  = {Choo, Jaegul and Lee, Changhyun and Reddy, Chandan K. and Park, Haesun},
  title   = {{UTOPIAN}: User-Driven Topic Modeling Based on Interactive Nonnegative Matrix Factorization},
  journal = {IEEE Transactions on Visualization and Computer Graphics},
  volume  = {19},
  number  = {12},
  pages   = {1992--2001},
  year    = {2013},
  doi     = {10.1109/TVCG.2013.212}
}

@article{park2017conceptvector,
  author  = {Park, Daejin and Kim, Seulgi and Lee, Joonhwan and Choo, Jaegul and Diakopoulos, Nicholas and Elmqvist, Niklas},
  title   = {{ConceptVector}: Text Visual Analytics via Interactive Lexicon Building Using Word Embedding},
  journal = {IEEE Transactions on Visualization and Computer Graphics},
  volume  = {24},
  number  = {1},
  pages   = {361--370},
  year    = {2018},
  doi     = {10.1109/TVCG.2017.2744478}
}

@article{elassady2019semantic,
  author  = {El-Assady, Mennatallah and Kehlbeck, Rebecca and Collins, Christopher and Keim, Daniel A. and Deussen, Oliver},
  title   = {Semantic Concept Spaces: Guided Topic Model Refinement Using Word-Embedding Projections},
  journal = {IEEE Transactions on Visualization and Computer Graphics},
  volume  = {26},
  number  = {1},
  pages   = {1001--1011},
  year    = {2020},
  doi     = {10.1109/TVCG.2019.2934665}
}

@inproceedings{faruqui2015retrofitting,
  author    = {Faruqui, Manaal and Dodge, Jesse and Jauhar, Sujay Kumar and Dyer, Chris and Hovy, Eduard and Smith, Noah A.},
  title     = {Retrofitting Word Vectors to Semantic Lexicons},
  booktitle = {Proceedings of the 2015 Conference of the North American Chapter of the Association for Computational Linguistics: Human Language Technologies},
  pages     = {1606--1615},
  year      = {2015},
  doi       = {10.3115/v1/N15-1184}
}

@article{meinecke2021explaining,
  author  = {Meinecke, Christian and Wrisley, David J. and J{\"a}nicke, Stefan},
  title   = {Explaining Semi-Supervised Text Alignment Through Visualization},
  journal = {IEEE Transactions on Visualization and Computer Graphics},
  volume  = {28},
  number  = {12},
  pages   = {4797--4809},
  year    = {2022},
  doi     = {10.1109/TVCG.2021.3105899}
}

@article{huang2023va,
  author  = {Huang, Zeyang and Witschard, Daniel and Kucher, Kostiantyn and Kerren, Andreas},
  title   = {{VA} + Embeddings {STAR}: A State-of-the-Art Report on the Use of Embeddings in Visual Analytics},
  journal = {Computer Graphics Forum},
  volume  = {42},
  number  = {3},
  pages   = {539--571},
  year    = {2023},
  doi     = {10.1111/cgf.14859}
}

\clearpage

\appendix
\begin{center}

{\LARGE\bfseries Supplemental Material}\\[0.5em]

{\large KeySI: An Interaction Framework for Tuning Text Embeddings Based on Human Feedback}\\[0.5em]

Yan Zhu, Y. Chen, and Rebecca Faust

\end{center}

\vspace{1em}
\section{Additional User Study Materials}
\label{sec:additional_user_study_materials}
This section reports participant-level questionnaire responses and refinement selections from the user study. (See Tables~\ref{tab:raw_taskA}, \ref{tab:raw_taskA_preference}, \ref{tab:raw_taskB}, \ref{tab:raw_taskB_goals}, and \ref{tab:participant_background}).

\section{Scope of the Task A Comparison}
\label{sec:scope_taskA_comparison}
This section summarizes the main workflow differences between DeepSI and KeySI relevant to the interpretation of Task A (See Table~\ref{tab:comparison_scope}).

\section{Additional Quantitative Results}
\label{sec:additional_quantitative_results}
This section reports the full local-purity and concept-purity results across datasets and neighborhood sizes. Here, $n$ denotes the number of documents in the corresponding evaluation set. Overall, tuning consistently improves both local and concept purity across datasets, with the largest gains typically appearing in local purity (See Tables~\ref{tab:appendix_local_purity}, \ref{tab:appendix_concept_purity}, and \ref{tab:appendix_participant_level_results}).

\section{Runtime and Scalability}
\label{appendix:runtime}
Table~\ref{tab:runtime} reports example runtimes of the KeySI interaction pipeline on the datasets used in this work. Runtime includes keyword-group specification, feedback translation, model tuning, and visualization updates. All experiments were conducted on a machine equipped with an AMD Ryzen 9 9800X3D CPU and an NVIDIA RTX 5090 GPU.
These results are intended to provide a reference for the responsiveness of the current prototype rather than a comprehensive scalability evaluation. Runtime depends on corpus size, the number of retrieved documents, visualization updates, and the amount of user interaction. A formal evaluation of scalability and performance on larger corpora and more resource-constrained hardware remains future work.

\section{Details and Empirical Behavior of the Two-Stage Semantic Denoising Strategy}
\label{appendix:staged-denoising}
To balance pseudo-supervision purity and coverage, KeySI uses a two-stage semantic denoising strategy rather than relying on a single fixed denoising pass.

\paragraph{Stage 1: High-confidence seed denoising.}
Starting from the candidate documents retrieved by lexical matching, KeySI first applies a stricter denoising criterion to construct a high-confidence seed set for each concept group. This stage is designed to favor precision over coverage: only candidates with a clearer preference for one concept group over competing groups are retained. The resulting seed set provides a relatively clean supervision signal for the initial tuning step.

\paragraph{Intermediate update.}
KeySI then performs a short model update using the Stage~1 pseudo-supervision set. The purpose of this update is not to fully optimize the embedding model, but to partially improve the local organization of the embedding space so that concept-consistent documents and boundary cases can be better distinguished in the next denoising step.

\paragraph{Stage 2: Relaxed recovery denoising.}
After the intermediate update, KeySI recomputes document embeddings and prototype affinities, and performs a second denoising pass with a more permissive criterion. This second stage is intended to recover additional documents that may have been too ambiguous in the original pretrained space but become more consistent with the target concepts after the initial update. The resulting pseudo-supervision set is therefore broader than the Stage~1 seed set while remaining more selective than lexical retrieval alone.

\paragraph{Thresholding scheme.}
Both stages use the same prototype-affinity gap score defined in the main paper, but with different operating strictness. In practice, Stage~1 uses a more conservative threshold to prioritize seed purity, whereas Stage~2 uses a more permissive threshold to improve coverage after the embedding space has been partially aligned. We use adaptive, group-specific thresholds estimated from the gap-score distributions of candidates that currently prefer each concept group, so that denoising remains sensitive to differences in concept difficulty and candidate-set composition across groups.

\paragraph{Empirical behavior.}
We examined this staged design on multiple keyword-group settings, including stress-test cases in which off-topic keywords were intentionally injected into concept groups (See Table~\ref{tab:staged_denoising_detailed}). Across these tested settings, the two-stage strategy exhibited the intended trade-off behavior: compared with lexical retrieval alone, Stage~1 generally produced a cleaner but smaller pseudo-supervision set, while Stage~2 often recovered additional concept-consistent documents and improved coverage relative to Stage~1. The magnitude of this benefit varied across settings. In some cases, Stage~2 provided a noticeable purity--coverage improvement over Stage~1 alone; in others, the gain was modest, especially when the initial retrieval was already relatively clean. These observations suggest that the value of the staged design is not that it guarantees improvement in every case, but that it provides a practical mechanism for balancing pseudo-supervision precision and coverage under varying concept conditions.

\paragraph{Interpretation.}
These results support the use of staged semantic denoising as a pragmatic design choice for pseudo-supervision construction. A single strict pass may discard too many near-boundary yet useful documents, whereas a single permissive pass may admit too much noise. By first establishing a high-confidence seed set and then expanding it after a partial model update, KeySI can more flexibly trade off stability and coverage during concept-oriented embedding adaptation.

\section{Additional Usage Scenario: AG News}
\label{appendix:AGNEWS}
\begin{figure*}[h]
    \centering
    \includegraphics[width=.9\linewidth]{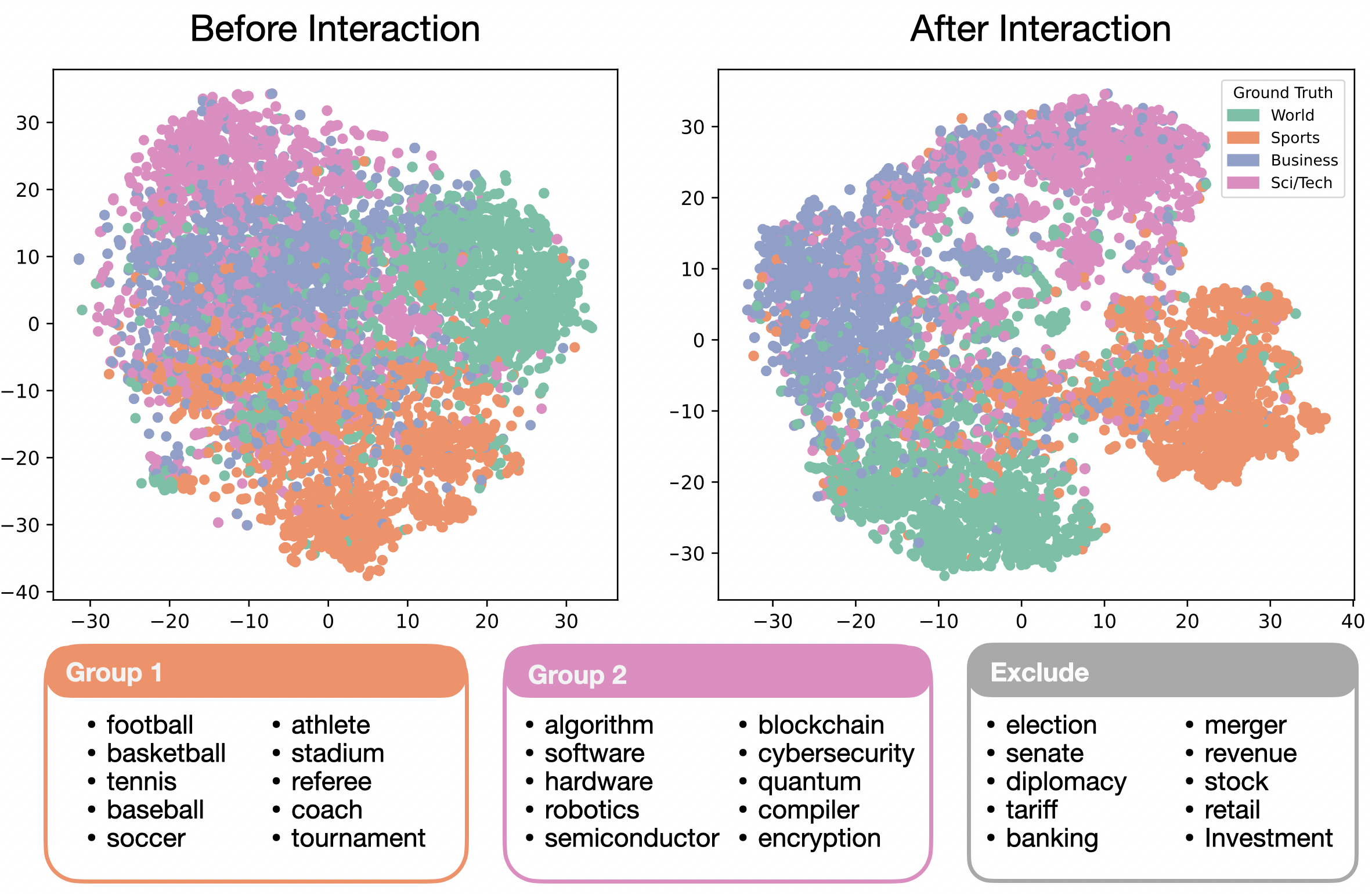}
    \caption{\textbf{AGNews Dataset.}}
    \label{fig:ag}
\end{figure*}

We tested \systemname{} on the AG News data set ($n=7{,}600$). In this example, we focused on two concepts, \emph{Sports} and \emph{Sci/Tech}. We selected ten keywords for each concept. Based on these keyword groups, \systemname{} retrieved candidate documents and, after semantic denoising, kept 120 documents for tuning.  The results are shown in Fig.~\ref{fig:ag}. Even though this filtered set is small compared with the full corpus, it still improved the overall embedding structure. Before tuning, the \emph{Sports} and \emph{Sci/Tech} documents showed only coarse grouping and were not clearly separated. After tuning, documents from the two target concepts formed cleaner neighborhoods and the boundary between them became clearer. This is consistent with the quantitative results in Tables~\ref{tab:appendix_local_purity} and~\ref{tab:appendix_concept_purity}, together with Table~\ref{tab:silhouette} in the main paper, which show improvements in both concept purity and silhouette.

This example suggests that, in our AG News setting, a relatively small amount of keyword-based supervision can still be sufficient to improve the broader embedding organization.

\section{Refinement Interface}
\label{appendix:refine}
\begin{figure*}[h]
    \centering
    \includegraphics[width=\linewidth]{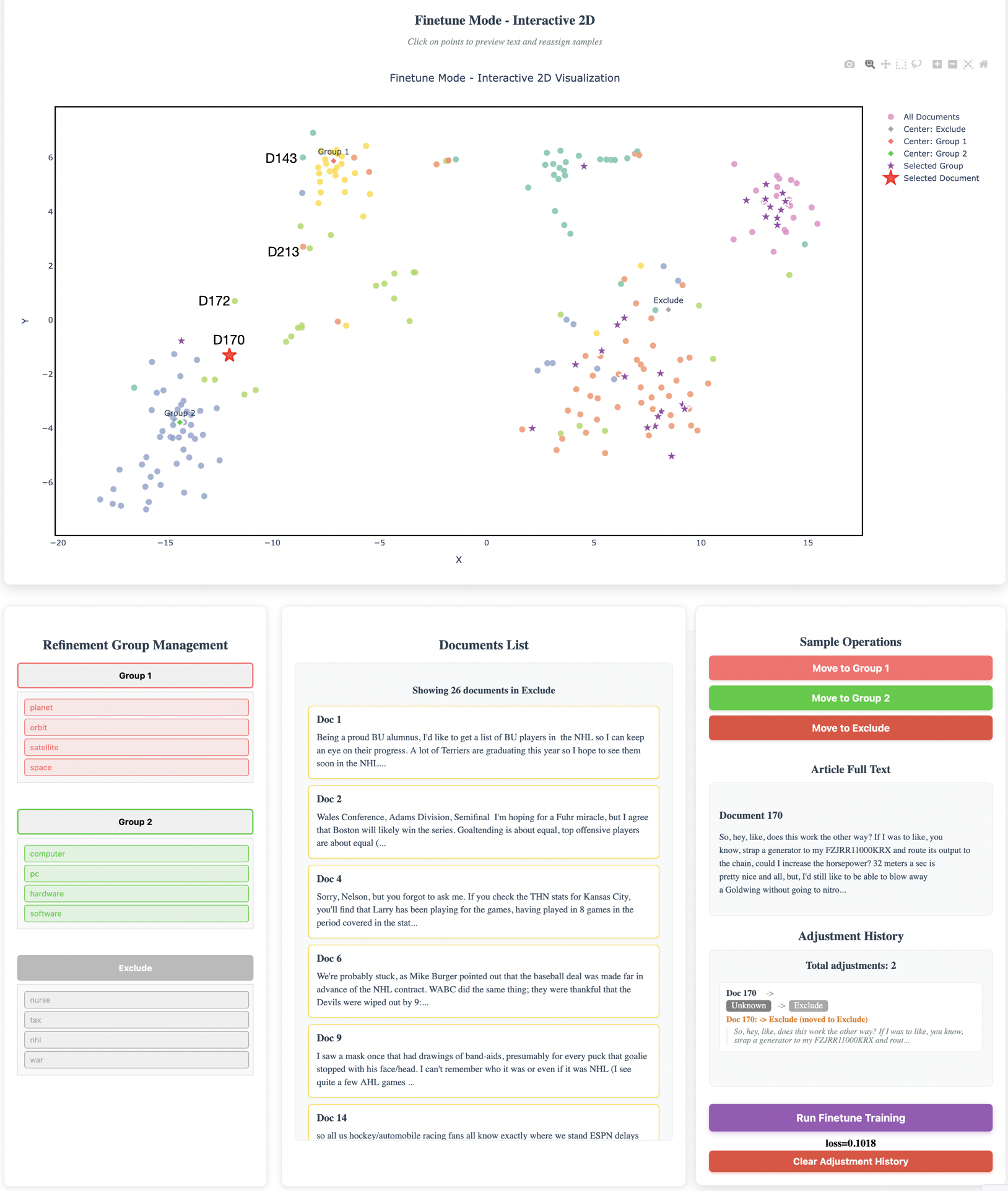}
    \caption{\textbf{Refinement interface.} Users can inspect documents in the projection to identify concept-irrelevant documents. The Refinement panel, in the bottom right, allows users to move documents to the groups or exclude them. The panel shows a preview of documents as well as a history of refinement actions. }
    \label{fig:refinementUI}
\end{figure*}

Figure~\ref{fig:refinementUI} illustrates the refinement interface. Users are able to inspect documents retrieved for each group as well as documents on the boundaries of the groups, to gauge concept clarity. The bottom right panel contains the refinement controls. For selected documents, users can inspect the contents to determine if it should remain in its current group (if it is in one), move to another group, or be excluded as off-topic. It also records a history of refinements to help users recall past actions. 

\begin{figure*}[th]
    \centering
    \includegraphics[width=.8\linewidth, trim={0cm 0cm 0cm 1.5cm},clip]{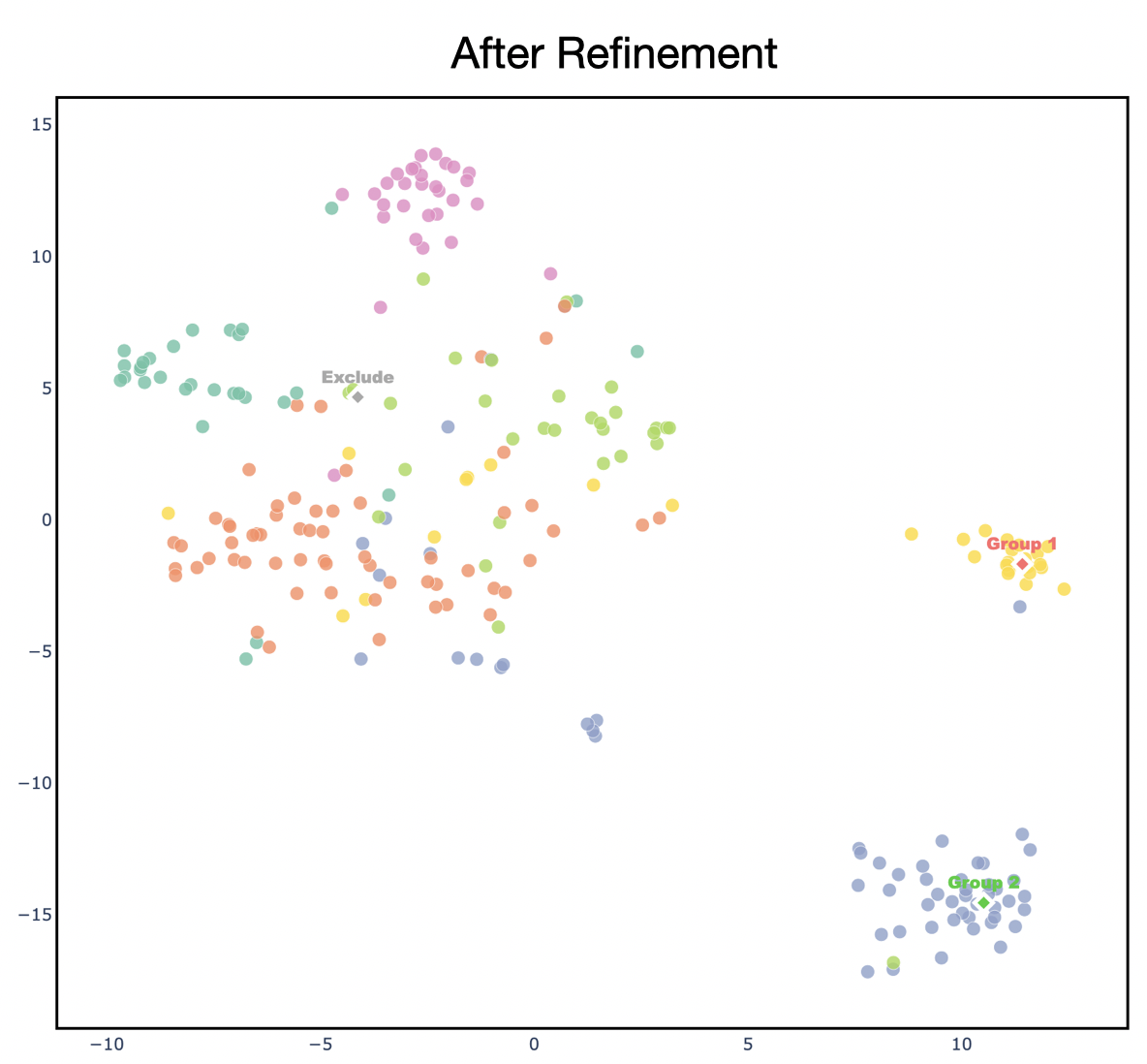}
    \caption{\textbf{Post Refinement Projection.} The embeddings after refinement. The boundaries become cleaner after refinement.}
    \label{fig:refined}
\end{figure*}

This example is a continuation of the 20 newsgroups usage scenario. The user identifies documents on the boundaries of the two concepts, as well as in between. They select four documents that are not relevant (those numbered in the projection) and move them all to the \texttt{Exclude} group. Figure~\ref{fig:refined} shows the embeddings after refinement. Notice, the boundaries are much cleaner after refinement. 

\section{Guidelines on Keyword Quantity and Quality}
\label{appendix:keyword-guidelines}
In \systemname{}, both the number and quality of keywords in each group directly affect the retrieved candidate documents, the purity of pseudo-supervision, and the final tuning outcome. In practice, we found that a small set of highly discriminative keywords often yields a more stable concept signal. Increasing the keyword count primarily improves coverage, but also increases the likelihood of ambiguity and noise.

\paragraph{Keyword count recommendations.}
For each concept group, we recommend starting with 3--5 keywords first to quickly establish a clear semantic direction. If coverage is insufficient (e.g., too few candidates or a sparse highlighted region), users can then expand the group incrementally. Adding many new keywords is more likely to introduce semantic drift (pulling in off-topic themes), with diminishing returns. For the \texttt{Exclude} group, we recommend 3--5 keywords to express a direction that is clearly out-of-scope for all target concepts. \texttt{Exclude} should avoid general terms; otherwise it can suppress a large portion of candidates and reduce system usability.

\paragraph{Keyword quality criteria.}
We suggest the following criteria when selecting keywords:
\begin{enumerate}
    \item \textbf{High concept specificity.} Prefer domain-specific terms, named entities, technical nouns, or abbreviations. Avoid common words that appear across multiple topics.
    \item \textbf{Low ambiguity.} Prioritize keywords whose meaning is stable within the dataset context. Terms such as player, student, or motorcycle may correspond to multiple topics and lead to mixed candidates.
    \item \textbf{Complementary coverage.} Keywords within a group should cover different facets of the concept (e.g., subtopics, entities, events) while maintaining a consistent semantic direction. Avoid adding many near-synonyms that provide little additional information.
    \item \textbf{High verifiability.} Keywords should allow users to reliably judge relevance via quick inspection (e.g., top-keyword previews or short snippets), before resorting to full-text reading.
\end{enumerate}

\paragraph{Practical selection workflow.}
We recommend an iterative workflow:
\begin{enumerate}
    \item \textbf{Start with the most specific anchor terms.} Add 1--2 highly representative keywords first and inspect whether retrieved candidates and projection highlights form a relatively concentrated region.
    \item \textbf{Use candidate previews for rapid quality checks.} Inspect the candidate list using top-keyword previews or short snippets (and open full text when needed) to estimate topic mixing. If candidates clearly span multiple themes or the keyword meaning shifts across contexts, replace that keyword.
    \item \textbf{Add keywords only to address two cases:}
    (i) \emph{insufficient coverage} (too few candidates / sparse regions), where adding more specific entities or terms within the same concept can improve recall; and
    (ii) \emph{boundary mixing} (candidates overlap with other concept regions), where users should replace ambiguous terms with more specific ones, or move noise-inducing directions into \texttt{Exclude}.
    \item \textbf{Iterate rather than aiming for a perfect initial definition.} In open-ended exploration, keyword groups typically converge from coarse to fine; users should refine groups progressively based on system feedback.
\end{enumerate}

\paragraph{Common pitfalls.}
We observed several recurring issues:
\begin{itemize}
    \item \textbf{Using generic words as concept keywords} (e.g., news, people, thing, good), which retrieves many irrelevant documents, shifts group prototypes, and makes gap-based denoising less reliable.
    \item \textbf{Using cross-topic terms as the concept core} (e.g., player), which can refer to sports players, game players, or even media players, leading to mixed pseudo-supervision and limited tuning gains.
    \item \textbf{Overly general \texttt{Exclude} terms}, which may suppress documents that should belong to target concepts, making the system appear ``hard to tune''.
\end{itemize}

\paragraph{How \systemname{} mitigates suboptimal keywords.}
Candidate previews and projection highlights provide \emph{online validation} of keyword quality. If a keyword consistently yields mixed candidates or produces highly dispersed highlights, it is typically overly broad or ambiguous and should be replaced or removed. If boundary mixing persists after tuning, \texttt{Refinement} serves as a fallback mechanism: users can locate suspicious points in the projection, verify them via preview/full-text inspection, reassign them to a better-matching group (or \texttt{Exclude}), and retrain to more precisely correct concept boundaries and scope.

\clearpage
\onecolumn

\begin{table}[!htbp]
\centering
\scriptsize
\setlength{\tabcolsep}{3pt}
\begin{adjustbox}{width=\textwidth}
\begin{tabular}{p{5.8cm}cccccccccccccccc}
\toprule
\multirow{2}{*}{\textbf{Item}} 
& \multicolumn{2}{c}{\textbf{P1}} 
& \multicolumn{2}{c}{\textbf{P2}} 
& \multicolumn{2}{c}{\textbf{P3}} 
& \multicolumn{2}{c}{\textbf{P4}} 
& \multicolumn{2}{c}{\textbf{P5}} 
& \multicolumn{2}{c}{\textbf{P6}} 
& \multicolumn{2}{c}{\textbf{P7}} 
& \multicolumn{2}{c}{\textbf{P8}} \\
\cmidrule(lr){2-3}\cmidrule(lr){4-5}\cmidrule(lr){6-7}\cmidrule(lr){8-9}
\cmidrule(lr){10-11}\cmidrule(lr){12-13}\cmidrule(lr){14-15}\cmidrule(lr){16-17}
& K & D & K & D & K & D & K & D & K & D & K & D & K & D & K & D \\
\midrule
Mental Demand -- How mentally demanding was the task? 
& 1 & 2 & 1 & 7 & 2 & 4 & 4 & 3 & 2 & 5 & 1 & 5 & 1 & 6 & 5 & 4 \\
Physical Demand -- How physically demanding was the task? 
& 1 & 2 & 1 & 1 & 1 & 2 & 2 & 4 & 1 & 3 & 1 & 5 & 1 & 1 & 1 & 1 \\
Temporal Demand -- How hurried or rushed was the pace of the task? 
& 1 & 2 & 1 & 6 & 1 & 3 & 2 & 3 & 3 & 6 & 1 & 5 & 1 & 5 & 1 & 3 \\
Performance -- How successful were you in accomplishing what you were asked to do? 
& 7 & 7 & 7 & 4 & 2 & 1 & 5 & 5 & 5 & 4 & 5 & 4 & 7 & 3 & 6 & 6 \\
Effort -- How hard did you have to work to accomplish your level of performance? 
& 1 & 2 & 1 & 6 & 1 & 1 & 5 & 5 & 4 & 5 & 1 & 5 & 1 & 5 & 3 & 3 \\
Frustration -- How insecure, discouraged, irritated, stressed, and annoyed were you? 
& 1 & 1 & 1 & 6 & 1 & 1 & 1 & 2 & 2 & 5 & 1 & 3 & 1 & 3 & 2 & 3 \\
\bottomrule
\end{tabular}
\end{adjustbox}
\caption{\textbf{Raw Task A questionnaire and workload responses.} K = KeySI; D = DeepSI. Items use their original 1--5 or 1--7 response scales.}
\label{tab:raw_taskA}
\end{table}

\begin{table}[!htbp]
\centering
\begin{tabular}{p{10cm}cccccccc}
\toprule
\textbf{Question} & \textbf{P1} & \textbf{P2} & \textbf{P3} & \textbf{P4} & \textbf{P5} & \textbf{P6} & \textbf{P7} & \textbf{P8} \\
\midrule
Which version of the interaction did you prefer overall? 
& K & K & K & K & K & K & K & K \\
Which version helped you better tune the model to reflect the target topics? 
& K & K & K & D & K & K & K & K \\
Which version felt more intuitive or natural to use? 
& K & K & K & K & K & K & K & K \\
Which version helped you better understand the topics in the document set? 
& D & K & D & K & D & D & D & K \\
Which version do you think would be more helpful in real-world tasks? 
& K & K & K & K & K & K & K & K \\
\bottomrule
\end{tabular}
\caption{\textbf{Raw Task A forced-choice preference responses.} K = KeySI; D = DeepSI.}
\label{tab:raw_taskA_preference}
\end{table}

\begin{table}[!htbp]
\centering
\scriptsize
\setlength{\tabcolsep}{5pt}
\begin{adjustbox}{width=\textwidth}
\begin{tabular}{p{10cm}cccccccc}
\toprule
\textbf{Item} & \textbf{P1} & \textbf{P2} & \textbf{P3} & \textbf{P4} & \textbf{P5} & \textbf{P6} & \textbf{P7} & \textbf{P8} \\
\midrule
I was able to form a clear semantic goal while using the interaction. 
& 6 & 7 & 5 & 6 & 6 & 7 & 6 & 6 \\
The keyword-based interaction helped me explore and understand the structure of the data. 
& 7 & 7 & 6 & 5 & 7 & 6 & 7 & 7 \\
The final result reflected my feedback and intentions. 
& 7 & 7 & 5 & 6 & 5 & 6 & 6 & 7 \\
I felt that I could influence how the model organized the documents. 
& 7 & 7 & 6 & 7 & 5 & 7 & 6 & 5 \\
The refinement view helped me express my semantic intent more precisely and reduce mixed documents. 
& 7 & 7 & 7 & 6 & 5 & 7 & 7 & 7 \\
\bottomrule
\end{tabular}
\end{adjustbox}
\caption{\textbf{Raw Task B questionnaire responses.} Items use their original questionnaire response scales.}
\label{tab:raw_taskB}
\end{table}

\begin{table}[!htbp]
\centering
\small
\setlength{\tabcolsep}{5pt}
\begin{tabular}{lcccc}
\toprule
\textbf{Participant} & \textbf{Reduce Mixed} & \textbf{Clearer Boundaries} & \textbf{Internal Consistency} & \textbf{Add More Docs} \\
\midrule
P1 & $\checkmark$ & $\checkmark$ &  &  \\
P2 & $\checkmark$ & $\checkmark$ & $\checkmark$ &  \\
P3 & $\checkmark$ &  & $\checkmark$ &  \\
P4 & $\checkmark$ & $\checkmark$ & $\checkmark$ & $\checkmark$ \\
P5 &  & $\checkmark$ & $\checkmark$ &  \\
P6 & $\checkmark$ & $\checkmark$ &  & $\checkmark$ \\
P7 & $\checkmark$ & $\checkmark$ & $\checkmark$ & $\checkmark$ \\
P8 &  &  & $\checkmark$ & $\checkmark$ \\
\bottomrule
\end{tabular}
\caption{\textbf{Participant-selected refinement goals in Task~B.}}
\label{tab:raw_taskB_goals}
\end{table}

\begin{table}[t]
\centering
\small
\begin{tabular}{lccc}
\toprule
\textbf{Participant} & \textbf{Background} & \textbf{ML Familiarity} & \textbf{ML Usage Frequency} \\
\midrule
P1 & CS PhD & 6 & 6 \\
P2 & CS PhD & 5 & 3 \\
P3 & CS PhD & 4 & 4 \\
P4 & CS PhD & 4 & 4 \\
P5 & Biology PhD & 1 & 1 \\
P6 & Business Master's & 1 & 1 \\
P7 & Non-CS / Professional & 1 & 1 \\
P8 & Neuroscience PhD & 1 & 1 \\
\bottomrule
\end{tabular}
\caption{\textbf{Participant background information.} ML familiarity and ML usage frequency were self-reported on 7-point Likert scales.}
\label{tab:participant_background}
\end{table}

\begin{table}[!htbp]
\centering
\small
\setlength{\tabcolsep}{4pt}
\begin{tabular}{p{3.6cm}p{4.5cm}p{4.5cm}}
\toprule
\textbf{Aspect} & \textbf{DeepSI} & \textbf{KeySI} \\
\midrule
Primary interaction entry & Document-level spatial manipulation & Keyword-level concept specification \\
Primary interaction object & Documents in the projection & Keywords, with optional document-level refinement \\
Need to inspect documents before giving supervision & Typically yes & Often reduced initially through keyword grouping and candidate previews \\
How supervision is formed & Spatially manipulated document groupings & Keyword groups translated into pseudo-supervision \\
Projection role & Direct manipulation workspace & Structural reference for documents and linked inspection \\
Additional candidate preview support & Limited by direct document inspection & Candidate list with keyword summaries and linked highlighting \\
Explicit out-of-scope mechanism & Not explicit at keyword level & \texttt{Exclude} group \\
Refinement mechanism & Continued document-level manipulation & Document reassignment after keyword-based tuning \\
\bottomrule
\end{tabular}
\caption{\textbf{Scope of the Task A comparison.} Task A contrasts two end-to-end interaction paradigms for text semantic steering rather than isolating a single matched backend component.}
\label{tab:comparison_scope}
\end{table}

\begin{table}[!htbp]
\centering
\scriptsize
\setlength{\tabcolsep}{5pt}
\begin{tabular}{p{4.4cm}c ccc ccc ccc}
\toprule
\multirow{2}{*}{\textbf{Dataset / pair}} & \multirow{2}{*}{$\mathbf{n}$}
& \multicolumn{3}{c}{\textbf{Local Pur@5}}
& \multicolumn{3}{c}{\textbf{Local Pur@10}}
& \multicolumn{3}{c}{\textbf{Local Pur@20}} \\
\cmidrule(lr){3-5}\cmidrule(lr){6-8}\cmidrule(lr){9-11}
& & Pre & Tuned & $\Delta$ & Pre & Tuned & $\Delta$ & Pre & Tuned & $\Delta$ \\
\midrule
20news: Sports vs Computer & 240 & 0.3273 & 0.9545 & 0.6272 & 0.2955 & 0.8500 & 0.5545 & 0.2227 & 0.5045 & 0.2818 \\
20news: Space vs Politics & 240 & 0.4111 & 0.9222 & 0.5111 & 0.3500 & 0.7444 & 0.3944 & 0.2861 & 0.4889 & 0.2028 \\
20news: Vehicles vs Med & 240 & 0.1636 & 0.7636 & 0.6000 & 0.1273 & 0.4727 & 0.3454 & 0.1000 & 0.2455 & 0.1455 \\
COVID: Cancer vs Smoking & 62 & 0.4600 & 0.9600 & 0.5000 & 0.3700 & 0.8200 & 0.4500 & 0.2625 & 0.4550 & 0.1925 \\
AG News test: class 2 vs 4 & 7600 & 0.1650 & 0.5150 & 0.3500 & 0.1225 & 0.4892 & 0.3667 & 0.1000 & 0.4275 & 0.3275 \\
\bottomrule
\end{tabular}
\caption{\textbf{Additional local-purity results across different neighborhood sizes.} Local purity evaluates neighborhood consistency around the directly supervised/user-set documents. Higher is better.}
\label{tab:appendix_local_purity}
\end{table}

\begin{table}[!htbp]
\centering
\scriptsize
\setlength{\tabcolsep}{5pt}
\begin{tabular}{p{4.4cm}c ccc ccc ccc}
\toprule
\multirow{2}{*}{\textbf{Dataset / pair}} & \multirow{2}{*}{$\mathbf{n}$}
& \multicolumn{3}{c}{\textbf{Concept Pur@5}}
& \multicolumn{3}{c}{\textbf{Concept Pur@10}}
& \multicolumn{3}{c}{\textbf{Concept Pur@20}} \\
\cmidrule(lr){3-5}\cmidrule(lr){6-8}\cmidrule(lr){9-11}
& & Pre & Tuned & $\Delta$ & Pre & Tuned & $\Delta$ & Pre & Tuned & $\Delta$ \\
\midrule
20news: Sports vs Computer & 240 & 0.6978 & 0.8356 & 0.1378 & 0.6567 & 0.8256 & 0.1689 & 0.5528 & 0.8033 & 0.2505 \\
20news: Space vs Politics & 240 & 0.6244 & 0.7844 & 0.1600 & 0.5611 & 0.7544 & 0.1933 & 0.4694 & 0.7228 & 0.2534 \\
20news: Vehicles vs Med & 240 & 0.5200 & 0.7700 & 0.2500 & 0.4167 & 0.7600 & 0.3433 & 0.3192 & 0.7133 & 0.3941 \\
COVID: Cancer vs Smoking & 62 & 0.4640 & 1.0000 & 0.5360 & 0.3920 & 1.0000 & 0.6080 & 0.2920 & 0.5840 & 0.2920 \\
AG News test: class 2 vs 4 & 7600 & 0.7001 & 0.7625 & 0.0624 & 0.6781 & 0.7600 & 0.0819 & 0.6545 & 0.7540 & 0.0995 \\
\bottomrule
\end{tabular}
\caption{\textbf{Additional concept-purity results across different neighborhood sizes.} Concept purity evaluates neighborhood consistency over the full ground-truth concept sets. Higher is better.}
\label{tab:appendix_concept_purity}
\end{table}

\begin{table}[!htbp]
\centering
\scriptsize
\setlength{\tabcolsep}{4.5pt}
\renewcommand{\arraystretch}{1.12}
\begin{tabular}{p{2.5cm}p{1.8cm}c ccc ccc ccc}
\toprule
\textbf{Pair} & \textbf{Eval.} & \textbf{User}
& \multicolumn{3}{c}{\textbf{@5}}
& \multicolumn{3}{c}{\textbf{@10}}
& \multicolumn{3}{c}{\textbf{@20}} \\
\cmidrule(lr){4-6}\cmidrule(lr){7-9}\cmidrule(lr){10-12}
& &
& Pre & Tuned & $\Delta$
& Pre & Tuned & $\Delta$
& Pre & Tuned & $\Delta$ \\
\midrule
\multirow{8}{*}{\shortstack{Sports vs\\Computer}}
& \multirow{4}{*}{Direct}
& P1 & 0.3727 & 0.8545 & 0.4818 & 0.2909 & 0.7364 & 0.4455 & 0.1932 & 0.5023 & 0.3091 \\
& & P2 & 0.4769 & 0.9615 & 0.4846 & 0.3538 & 0.9385 & 0.5847 & 0.2173 & 0.6038 & 0.3865 \\
& & P3 & 0.4667 & 1.0000 & 0.5333 & 0.3571 & 0.8143 & 0.4572 & 0.2452 & 0.5048 & 0.2596 \\
& & P4 & 0.2286 & 0.8143 & 0.5857 & 0.2000 & 0.5857 & 0.3857 & 0.1429 & 0.3643 & 0.2214 \\
\cmidrule(lr){2-12}
& \multirow{4}{*}{Concept}
& P1 & 0.6911 & 0.8533 & 0.1622 & 0.6356 & 0.8411 & 0.2055 & 0.5361 & 0.8239 & 0.2878 \\
& & P2 & 0.6800 & 0.7711 & 0.0911 & 0.6378 & 0.7622 & 0.1244 & 0.5367 & 0.7411 & 0.2044 \\
& & P3 & 0.7089 & 0.8222 & 0.1133 & 0.6422 & 0.7933 & 0.1511 & 0.5606 & 0.7467 & 0.1861 \\
& & P4 & 0.6733 & 0.8244 & 0.1511 & 0.6233 & 0.8156 & 0.1923 & 0.5583 & 0.7739 & 0.2156 \\
\midrule
\multirow{8}{*}{\shortstack{Politics vs\\Space}}
& \multirow{4}{*}{Direct}
& P1 & 0.2364 & 0.9364 & 0.7000 & 0.2091 & 0.7909 & 0.5818 & 0.1682 & 0.5114 & 0.3432 \\
& & P2 & 0.2357 & 0.9429 & 0.7072 & 0.2321 & 0.9036 & 0.6715 & 0.1946 & 0.6500 & 0.4554 \\
& & P3 & 0.2333 & 0.9778 & 0.7445 & 0.1722 & 0.7722 & 0.6000 & 0.1694 & 0.4000 & 0.2306 \\
& & P4 & 0.2522 & 1.0000 & 0.7478 & 0.2130 & 0.9652 & 0.7522 & 0.1674 & 0.5261 & 0.3587 \\
\cmidrule(lr){2-12}
& \multirow{4}{*}{Concept}
& P1 & 0.5489 & 0.6689 & 0.1200 & 0.4856 & 0.6344 & 0.1488 & 0.4178 & 0.5989 & 0.1811 \\
& & P2 & 0.6289 & 0.7711 & 0.1422 & 0.5900 & 0.7711 & 0.1811 & 0.4967 & 0.7861 & 0.2894 \\
& & P3 & 0.5956 & 0.7000 & 0.1044 & 0.5589 & 0.6922 & 0.1333 & 0.4789 & 0.6722 & 0.1933 \\
& & P4 & 0.5689 & 0.7022 & 0.1333 & 0.5400 & 0.6856 & 0.1456 & 0.4572 & 0.6856 & 0.2284 \\
\bottomrule
\end{tabular}
\caption{\textbf{Participant-level results for the two 20news concept pairs.} For each concept pair, we report participant-level results under direct/user-set evaluation and concept-restricted evaluation. Higher is better.}
\label{tab:appendix_participant_level_results}
\end{table}

\begin{table}[h]
    \centering
    \begin{tabular}{c|cccc}
        \toprule
        Dataset & \# Concept & \# Keywords & \# Docs & Time \\
         & Groups & per Group & Retrieved & (s) \\
        \midrule
        COVID-19 & 3 & 3 & 28 & 60 \\
        20 News & 3 & 3 & 43 & 69 \\
        AG News & 3 & 7 & 576 & 210 \\
        \bottomrule
    \end{tabular}
    \caption{\textbf{Example performance.} Runtime of the three interactions.}
    \label{tab:runtime}
\end{table}

\begin{table*}[!htbp]
\centering
\scriptsize
\setlength{\tabcolsep}{3.5pt}
\begin{tabular}{lcccccccccccc}
\toprule
\multirow{2}{*}{\textbf{Case}} &
\multicolumn{4}{c}{\textbf{lexical retrieval}} &
\multicolumn{4}{c}{\textbf{Stage 1 denoising}} &
\multicolumn{4}{c}{\textbf{Stage 1+2 denoising}} \\
\cmidrule(lr){2-5}\cmidrule(lr){6-9}\cmidrule(lr){10-13}
& Purity & Coverage & NRR & TPR
& Purity & Coverage & NRR & TPR
& Purity & Coverage & NRR & TPR \\
\midrule
P1 & 0.4275 & 1.0000 & 0.0000 & 1.0000 & 0.5609 & 0.7084 & 0.3750 & 0.8750 & 0.6056 & 0.5997 & 0.5568 & 0.8333 \\
P2 & 0.6340 & 1.0000 & 0.0000 & 1.0000 & 0.7425 & 0.7723 & 0.4333 & 0.8944 & 0.7500 & 0.8035 & 0.4333 & 0.9444 \\
P3 & 0.6111 & 1.0000 & 0.0000 & 1.0000 & 0.8055 & 0.7500 & 0.5000 & 0.9091 & 0.8334 & 0.6666 & 0.6429 & 0.8636 \\
P4 & 0.7143 & 1.0000 & 0.0000 & 1.0000 & 0.9231 & 0.4325 & 0.8333 & 0.5521 & 0.9231 & 0.6706 & 0.8333 & 0.8646 \\
P5 & 0.7540 & 1.0000 & 0.0000 & 1.0000 & 0.9500 & 0.3968 & 0.9375 & 0.5750 & 0.9252 & 0.7103 & 0.4375 & 0.9000 \\
\midrule
Avg. & 0.6282 & 1.0000 & 0.0000 & 1.0000 & 0.7964 & 0.6120 & 0.6158 & 0.7611 & 0.8075 & 0.6901 & 0.5808 & 0.8812 \\
\bottomrule
\end{tabular}
\caption{\textbf{Detailed results for the two-stage semantic denoising strategy under stress-test keyword settings.} Coverage is reported relative to the lexical retrieval candidate pool. TPR denotes true-positive retention. Stage~1 generally favors a cleaner seed set, while Stage~2 often recovers additional concept-consistent documents after the intermediate model update.}
\label{tab:staged_denoising_detailed}
\end{table*}

\end{document}